\documentclass{phai}

\usepackage{amsmath,amssymb}
\usepackage{tikz}
\usetikzlibrary{arrows.meta,positioning,fit,calc}
\usepackage{algorithm}
\usepackage{algpseudocode}
\usepackage{float}
\usepackage{url}

\hypersetup{pdfauthor={Dayou Li, Jionghao Wang, Peihao Wang, Jiuzhou Lei, Hao Wang, Jichen Chai, Hezhen Hu, Minghui Zheng, Zhangyang Wang, Xin Li, Wenping Wang, Zhiwen Fan},pdftitle={Disentangling Task-Relevant Latent Dynamics for Reward Modeling}}

\title{Disentangling Task-Relevant Latent Dynamics for Reward Modeling}
\author[1,*]{Dayou~Li}
\author[1,*]{Jionghao~Wang}
\author[2]{Peihao~Wang}
\author[1]{Jiuzhou~Lei}
\author[3]{Hao~Wang}
\author[4]{Jichen~Chai}
\author[2]{Hezhen~Hu}
\author[1]{Minghui~Zheng}
\author[2]{Zhangyang~Wang}
\author[1]{Xin~Li}
\author[1]{Wenping~Wang}
\author[1,\dagger]{Zhiwen~Fan}

\affiliation[1]{Texas A\&M University}
\affiliation[2]{University of Texas at Austin}
\affiliation[3]{Google DeepMind}
\affiliation[4]{Georgia Institute of Technology}
\contribution[*]{Equal contribution}
\contribution[\dagger]{Corresponding author}
\abstract{
Reward design remains a major bottleneck for reinforcement learning in robotic manipulation. Handcrafted dense rewards can compensate for the limited guidance provided by sparse rewards, but require substantial task-specific reward engineering. Learning rewards from videos provides a scalable, data-driven alternative to manual reward design. However, existing video-based rewards are often computed in entangled visual spaces where task-relevant state changes are confounded with extraneous visual variation. As a result, the learned reward may respond to visual distractors instead of actual task execution. To address this limitation, we propose \textbf{DiTaR} (\textbf{Di}sentangling \textbf{Ta}sk-Relevant Latent Dynamics for \textbf{R}eward Modeling), a framework for learning dense rewards from heterogeneous videos. 
DiTaR learns a continuous task-relevant representation from multi-frame video latents through an asymmetric two-stage procedure that suppresses task-irrelevant visual variation.
In the task-relevant latent space, we train a history- and language-conditioned rectified-flow model to capture task dynamics and derive a directional dense reward from temporal changes in dynamics compatibility. Together, these components enable reward modeling directly over task-relevant dynamics rather than generic visual variation. DiTaR substantially improves policy learning on MetaWorld-v3 and ManiSkill over video-based reward baselines, and transfers to real-world offline RL through reward relabeling. Reward analyses further show that DiTaR provides more reliable rewards.

}

\date{September 27, 2026}
\metadata[Keywords]{Reward Model, Robotic Manipulation}
\metadata[Project Website]{\url{https://ditarr.github.io/}}
\correspondence{Zhiwen Fan: \email{zhiwenfan@tamu.edu}}

\begin{document}
\maketitle

\section{Introduction}

Reward design remains a major bottleneck for reinforcement learning in robotic manipulation.
While sparse rewards are easy to specify, their binary signals provide limited guidance for efficient exploration, often leading to slow convergence or failed policy learning in long-horizon tasks.
Handcrafted dense rewards can accelerate optimization by providing continuous feedback, but typically require task-specific design using privileged simulator states, making them difficult to scale across diverse tasks.
Learning rewards from videos offers a scalable, data-driven alternative to such manual reward design~\citep{huang2024diffusionreward,zhang2025rewind,tan2025robodopamine,wang2025genreward,lee2026roboreward,zhai2025vlac,chen2025sarm}.

Videos provide a natural source of dense rewards because they record continuous state changes throughout task execution.
Prior work leverages this visual signal through video prediction likelihoods~\citep{escontrela2023viper}, history-conditioned diffusion uncertainty~\citep{huang2024diffusionreward}, or progress estimation~\citep{zhang2025rewind}.
These methods differ in how they derive rewards from videos, but their reward representations are not explicitly structured to separate task-relevant state changes from extraneous visual variation.
As a result, rewards computed from raw observations or generic visual features can respond to camera shifts, lighting changes, or background distractors rather than the state changes that matter for task completion.
An RL agent trained on such signals may therefore receive misleading feedback or reinforce correlations unrelated to the intended behavior.
This motivates two key requirements for video-based reward learning: evaluating transitions in a task-relevant representation space and providing directional feedback that reflects task performance.

Generic reconstruction and prediction objectives alone do not explicitly isolate the state changes that matter for task execution.
Recent work on vision-language-action (VLA) models similarly introduces task-centric latent actions and task-relevant factor specialization to focus representations on information relevant to task execution~\citep{bu2025univla,zhang2026clap,jia2026guidedvlaspecifyingtaskrelevantfactors}.
This issue is especially important for reinforcement learning because errors in the reward representation directly become optimization targets: if task-irrelevant visual variation influences the reward, policy optimization may reinforce these correlations instead of the intended behavior.
The latent space used for reward modeling should therefore retain state changes that support task completion while suppressing task-irrelevant visual variation.

A task-relevant representation alone is insufficient for reward construction.
Likelihood- or uncertainty-based video rewards measure how well an observation agrees with the learned distribution, but an absolute measure does not necessarily indicate whether the execution is improving relative to its recent history.
Progress-based approaches such as ReWiND~\citep{zhang2025rewind} address directionality by directly predicting task progress, but obtain supervision from predefined progress targets and temporally transformed demonstrations, while the progress estimator operates on generic pretrained visual features.
This motivates deriving directional rewards directly from learned task-relevant dynamics without relying on explicit progress supervision from temporal transformations.

\begin{figure*}[t]
    \centering
    \includegraphics[width=\textwidth]{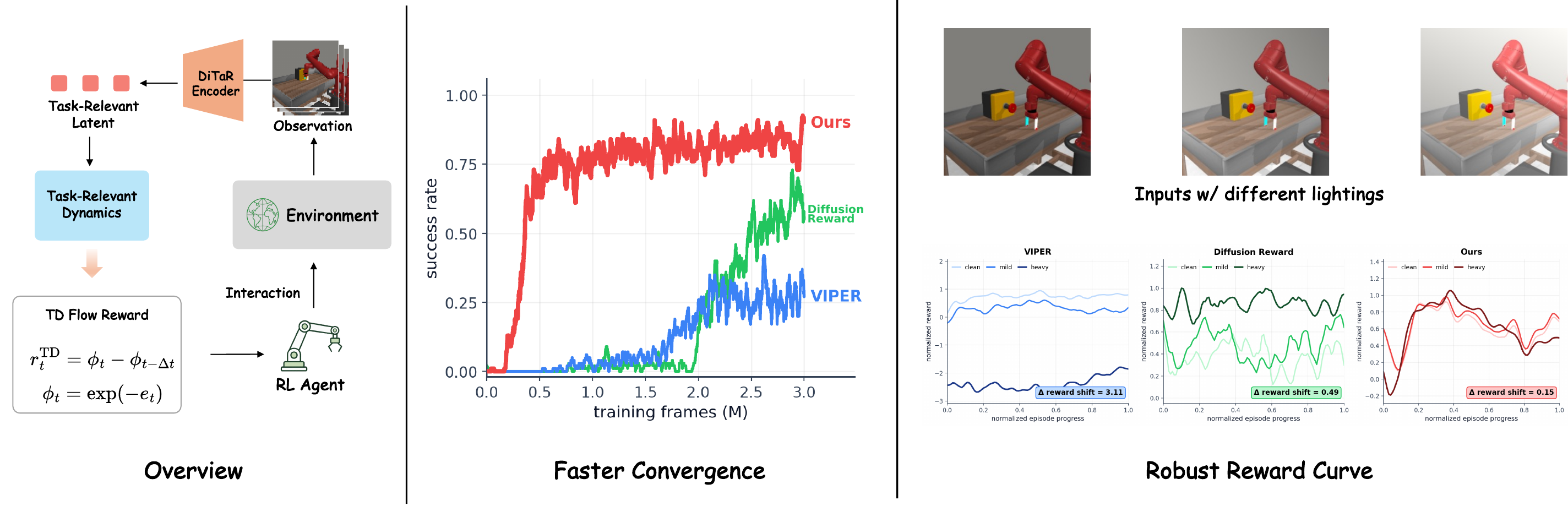}
    \caption{
    DiTaR learns dense rewards from task-relevant latent dynamics.
    The DiTaR reward accelerates RL convergence while remaining stable under nuisance visual variations, enabling robust policy learning across diverse tasks.
    }
    \label{fig:teaser}

\end{figure*}

To satisfy these requirements, we propose \textbf{DiTaR} (shown in Figure~\ref{fig:teaser}), a framework that learns dense rewards from videos by modeling task-relevant latent dynamics.
DiTaR first uses an asymmetric two-stage representation learning procedure over multi-frame video latents to obtain a continuous task-relevant representation while separating task-irrelevant visual variation.
In the first stage, language explicitly provides task semantics, encouraging a capacity-limited latent to encode complementary visual information beyond the language signal.
In the second stage, we freeze the first-stage pathway and remove the decoder's direct language conditioning, so that a separate latent must recover from video the task-relevant information previously supplied by language.
DiTaR then models task dynamics in this task-relevant latent space with a history- and language-conditioned rectified-flow model.
A directional dense reward is derived from how the agreement between observed transitions and the learned task dynamics changes over time, which is evaluated for policy learning on MetaWorld-v3 and ManiSkill.

Our main contributions are summarized as follows:
\begin{itemize}
\item We introduce a continuous task-relevant representation learning framework over multi-frame video latents, using an asymmetric two-stage representation learning procedure to yield a task-relevant latent space for reward modeling from action-free videos.
\item We model task dynamics in this latent space with a history- and language-conditioned rectified-flow model, and derive a directional dense reward from temporal changes in how observed transitions align with the learned dynamics, without explicit progress supervision.
\item We evaluate DiTaR for visual RL on MetaWorldv3 and ManiSkill, and further validate it in real-world offline RL by relabeling robot demonstrations with DiTaR rewards. Reward and representation analyses further demonstrate robustness to nuisance visual variation and sensitivity to task-relevant temporal structure.
\end{itemize}

\section{Related Work}

\noindent \textbf{Learning Rewards from Videos.}
Prior work learns dense rewards from videos through visual alignment, generative modeling, or progress estimation~\citep{aytar2018playing,sermanet2018time,dwibedi2019temporal,chen2021learning,alakuijala2023learning,yang2024rank2reward}.
VIP and LIV learn reward-oriented visual representations, while RoboCLIP, VLM-RM, Adapt2Reward, GVL, and VICtoR use visual or vision-language alignment~\citep{ma2023vip,ma2023liv,sontakke2023roboclip,rocamonde2024vlmrm,yang2024adapt2reward,ma2025vision,hung2025victor}.
Generative approaches derive rewards from video likelihood, uncertainty, or generated goals~\citep{escontrela2023viper,huang2024diffusionreward,luo2024text,wang2025genreward}.
Recent methods also learn progress-, goal-, or subtask-based rewards, but often operate on raw observations, generated-goal similarity, or generic visual features without explicitly separating task-relevant dynamics from irrelevant visual variation~\citep{alakuijala2024video,zhang2025rewind,lee2026roboreward,zhai2025vlac,chen2025sarm,tan2025robodopamine,biza2025robot,ayalew2025progressor,kim2025subtask,liu2025timerewarder}.

\noindent \textbf{Task-Relevant Representation.}
Robotic representation learning has shifted from generic visual features to spaces that capture action-relevant state change.
General and self-supervised pretraining methods provide robust features for robot learning and reward specification~\citep{nair2023r3m,ma2023vip,karamcheti2023language,oquab2024dinov2,assran2023self,bardesrevisiting}.
Recent work further targets controllable structure by learning latent actions~\citep{ye2025latent}, including UniVLA~\citep{bu2025univla}, which learns vector-quantized task-centric latent actions from paired-frame DINOv2 features for policy learning; other approaches filter task-irrelevant factors~\citep{zhang2026clap}, ground task factors in VLA policies~\citep{jia2026guidedvlaspecifyingtaskrelevantfactors}, or extract controllable changes through video generation, action grounding, and point tracks~\citep{du2023learning,luo2025grounding,bharadhwaj2024track2act}.
Latent-dynamics and world-model methods capture temporal evolution by predicting future latent states or features~\citep{hafner2023mastering,hansen2024td,bruce2024genie,zhou2025dinowm,assran2025v,mur2026v,maes2026leworldmodel,lyu2026lda}, motivating our use of task-relevant latent dynamics for video-based reward learning.

\section{Methodology}
\begin{figure}[t]
    \centering
    \includegraphics[width=\linewidth]{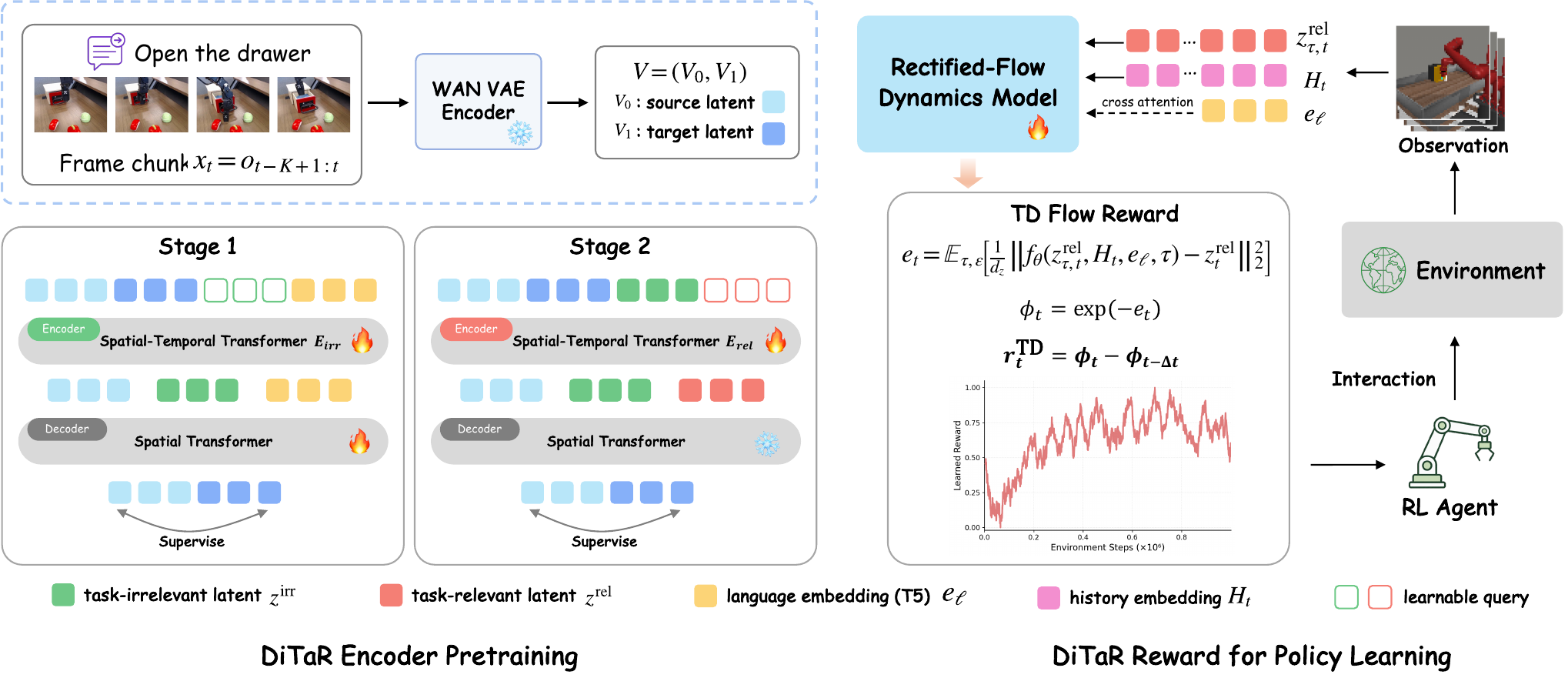}
    \caption{
    \textbf{Pipeline overview.}
    DiTaR disentangles task-irrelevant and task-relevant latents through an asymmetric two-stage representation learning procedure.
    A history- and language-conditioned rectified-flow model captures task-relevant transition dynamics, and the temporal difference of its residual-based compatibility score provides the dense reward.
    }
    \label{fig:pipeline}

\end{figure}

As shown in Figure~\ref{fig:pipeline}, DiTaR learns a dense reward from action-free videos by modeling task-relevant temporal evolution in a continuous multi-frame video latent space.
It first uses an asymmetric two-stage representation learning procedure with KL-regularized bottlenecks to separate task-relevant state changes from task-irrelevant visual variation (Sec.~\ref{sec:latent_disentangle}).
DiTaR then models temporal dynamics directly in this task-relevant latent space, learning the conditional distribution of the target latent given recent visual history and task instruction (Sec.~\ref{sec:task_dynamics}).
The resulting residual measures how well an observed transition agrees with the learned task-conditioned dynamics, and its temporal change provides a directional dense reward for policy learning (Sec.~\ref{sec:reward_construction}).
Implementation details and pseudocode are provided in Appendices~\ref{sec:suppl_impl} and~\ref{sec:suppl_algorithms}.

\subsection{Problem Formulation}

We consider an RL agent interacting with a partially observable environment using only RGB observations.
At time $t$, the agent observes $o_t\in\mathcal{O}$ and samples an action
$a_t\sim\pi_\omega(\cdot\mid o_t)$.
Each task is specified by a language instruction $\ell$.
We assume an action-free video dataset
$\mathcal{D}=\{(\xi^i,\ell^i)\}_{i=1}^{N}$, where
$\xi^i=(o_0^i,\ldots,o_{T_i}^i)$ is a task-execution video paired with its instruction $\ell^i$.
The videos provide visual observations and language instructions without action labels or privileged environment states.

For reward evaluation at time $t$, we define the target clip $x_t$ and its preceding visual history $h_t$ as
\begin{equation}
x_t
=
(o_{t-K+1},\ldots,o_t),
\qquad
h_t
=
(o_{t-K-H+1},\ldots,o_{t-K}),
\label{eq:history_target}
\end{equation}
where $K$ is the target-clip length and $H$ is the history length.
Both $x_t$ and $h_t$ contain only observations available up to time $t$.
Let $e_\ell=f_{\mathrm{text}}(\ell)$ denote the language embedding.

DiTaR consists of three learned components.
First, the two-stage latent encoders $E_{\mathrm{irr}}$ and $E_{\mathrm{rel}}$ learn a task-relevant latent $z_t^{\mathrm{rel}}$ from the target clip while separating task-irrelevant visual variation.
Second, a conditional dynamics model
$p_\theta(z_t^{\mathrm{rel}}\mid H_t,e_\ell)$,
with $H_t=E_{\mathrm{Wan}}(h_t)$,
models task-relevant dynamics from the preceding visual history and task instruction.
Finally, the resulting dynamics residual is converted into a dense reward $r_t^{\mathrm{TD}}$ for policy learning.

\subsection{Task-Relevant Latent Encoding}
\label{sec:latent_disentangle}

DiTaR learns continuous task-relevant representations from multi-frame video latents through an asymmetric two-stage procedure. In Stage 1, language supplies task semantics, encouraging a capacity-limited latent to capture complementary visual information for reconstruction. In Stage 2, we freeze this pathway and remove language conditioning from the decoder, so that a second latent must recover the task-relevant information directly from video.
\paragraph{Base Latent Space.}
We instantiate DiTaR on the latent space of a frozen Wan VAE~\citep{wan2025wan}.
Given a strided 5-frame clip $x$, the encoder produces source and future video latent tokens
$V=(V_0,V_1)=E_{\mathrm{Wan}}(x)$,
where $V_0$ and $V_1$ denote the source and future latents, respectively.
DiTaR operates on these base latents using two spatial-temporal Transformer encoders, $E_{\mathrm{irr}}$ and $E_{\mathrm{rel}}$, each equipped with separate learnable query tokens and a KL-regularized variational bottleneck.

\paragraph{Stage 1: Task-Irrelevant Latent.}
The language embedding $e_\ell$ is obtained from a frozen T5 encoder~\citep{raffel2020exploring}.
Stage~1 trains $E_{\mathrm{irr}}$ together with a reconstruction decoder $D_\psi$, while making $e_\ell$ explicitly available during both encoding and reconstruction.
Using learnable queries $Q_{\mathrm{irr}}$, the encoder produces
\begin{equation}
z^{\mathrm{irr}}
\sim
q_{\eta_{\mathrm{irr}}}
\left(
z^{\mathrm{irr}}
\mid
V,e_\ell
\right).
\end{equation}
The decoder reconstructs the future-target video latent as
\begin{equation}
\hat V_1^{(1)}
=
V_0+
D_\psi
\left(
V_0,z^{\mathrm{irr}},e_\ell
\right).
\label{eq:stage1_reconstruction}
\end{equation}
Because task-level semantics are already supplied through the language pathway, the capacity-limited $z^{\mathrm{irr}}$ is encouraged to retain complementary visual information required to reconstruct $V_1$.

\paragraph{Stage 2: Task-Relevant Latent.}
Stage~2 removes the decoder's direct access to the language embedding while preserving the representation learned in Stage~1.
We freeze both the Stage~1 encoder $E_{\mathrm{irr}}$ and reconstruction decoder $D_\psi$, and introduce a second encoder $E_{\mathrm{rel}}$ with learnable queries $Q_{\mathrm{rel}}$.
The frozen Stage~1 encoder provides $z^{\mathrm{irr}}$, while $E_{\mathrm{rel}}$ produces
\begin{equation}
z^{\mathrm{rel}}
\sim
q_{\eta_{\mathrm{rel}}}
\left(
z^{\mathrm{rel}}
\mid
V,z^{\mathrm{irr}}
\right).
\end{equation}
The fixed decoder then reconstructs the future-target latent using $z^{\mathrm{rel}}$ in place of the explicit language conditioning used in Stage~1:
\begin{equation}
\hat V_1^{(2)}
=
V_0+
D_\psi
\left(
V_0,
z^{\mathrm{irr}},
z^{\mathrm{rel}}
\right).
\label{eq:stage2_reconstruction}
\end{equation}
Because $E_{\mathrm{irr}}$ and $D_\psi$ remain fixed, $E_{\mathrm{rel}}$ is the only optimized component in Stage~2.
It must recover the task-relevant information that was previously supplied directly through language.

Both stages use the same KL-regularized future-target reconstruction objective:
\begin{equation}
\begin{aligned}
\mathcal{L}_s
&=
\mathcal{L}_{\mathrm{rec}}^s
+
\beta_s\mathcal{L}_{\mathrm{KL}}^s,
\qquad
s\in\{\mathrm{irr},\mathrm{rel}\},
\\
\mathcal{L}_{\mathrm{rec}}^s
&=
\mathbb{E}_{z^s\sim q_s}
\left[
\frac{1}{d_V}
\left\|
\hat V_1^{(s)}-V_1
\right\|_2^2
\right],
\qquad
\mathcal{L}_{\mathrm{KL}}^s
=
D_{\mathrm{KL}}
\left(
q_s
\,\Vert\,
\mathcal{N}(0,I)
\right),
\end{aligned}
\label{eq:latent_objective}
\end{equation}
where $q_s$ denotes the corresponding variational posterior, $d_V$ denotes the dimensionality of the flattened target latent, and $\beta_s$ controls the strength of the KL regularization.
\subsection{Conditional Rectified-Flow Dynamics}
\label{sec:task_dynamics}

The encoder maps the observed clip $x_t$ into a task-relevant latent representation but does not model how this representation evolves over time.
We therefore model the conditional distribution $p_\theta(z_t^{\mathrm{rel}}\mid H_t,e_\ell)$, where the visual history is encoded by the frozen Wan VAE as $H_t=E_{\mathrm{Wan}}(h_t)$.
This conditional distribution captures task-relevant dynamics given the history and task instruction.

We model this distribution with rectified flow~\citep{esser2024scaling}.
Given the task-relevant latent $z_t^{\mathrm{rel}}$ and Gaussian noise $\epsilon\sim\mathcal{N}(0,I)$, we construct the linear probability path
\begin{equation}
z_{\tau,t}
=
(1-\tau)z_t^{\mathrm{rel}}
+
\tau\epsilon,
\qquad
\tau\sim\mathcal{U}(\tau_{\min},1),
\label{eq:flow_path}
\end{equation}
where $0<\tau_{\min}<1$, with $\tau=0$ corresponding to the data endpoint and $\tau=1$ to the noise endpoint.

Instead of directly predicting the flow velocity, DiTaR parameterizes the rectified-flow model by predicting the data endpoint, providing a direct discrepancy in the task-relevant latent space for subsequent reward construction:
\begin{equation}
\hat z_{0,t}
=
f_\theta
\left(
z_{\tau,t},
H_t,
e_\ell,
\tau
\right).
\label{eq:endpoint_prediction}
\end{equation}
Under Eq.~\eqref{eq:flow_path}, this endpoint parameterization induces the same rectified-flow field as velocity prediction, with endpoint regression corresponding to a $\tau^2$-weighted velocity-regression objective.

We train the dynamics model with
\begin{equation}
\mathcal{L}_{\mathrm{end}}
=
\mathbb{E}_{t,\tau,\epsilon}
\left[
\frac{1}{d_z}
\left\|
\hat z_{0,t}
-
z_t^{\mathrm{rel}}
\right\|_2^2
\right],
\label{eq:endpoint_objective}
\end{equation}
where $d_z$ denotes the dimensionality of the task-relevant latent.
The resulting endpoint discrepancy measures the deviation between the observed task-relevant latent and the endpoint predicted under the learned conditional distribution, and serves as the residual for reward construction in Sec.~\ref{sec:reward_construction}.
We instantiate $f_\theta$ with an MMDiT~\citep{esser2024scaling} backbone.

\subsection{Temporal-Difference Flow Reward}
\label{sec:reward_construction}

The endpoint residual provides a measure of how compatible the observed task-relevant latent is with the learned task-relevant dynamics under the visual history and task instruction.
Given the observed task-relevant latent $z_t^{\mathrm{rel}}$, we sample $\epsilon\sim\mathcal{N}(0,I)$ and $\tau\sim\mathcal{U}(\tau_{\min},\tau_{\max})$, where $0<\tau_{\min}<\tau_{\max}\leq1$, and construct $z_{\tau,t}$ using Eq.~\eqref{eq:flow_path}. We define the endpoint residual as
\begin{equation}
e_t
=
\mathbb{E}_{\tau,\epsilon}
\left[
\frac{1}{d_z}
\left\|
f_\theta
\left(
z_{\tau,t},
H_t,
e_\ell,
\tau
\right)
-
z_t^{\mathrm{rel}}
\right\|_2^2
\right].
\label{eq:transition_residual}
\end{equation}
A smaller $e_t$ indicates greater compatibility between the observed transition and the learned task-conditioned dynamics.

We convert the residual into a bounded compatibility score $\phi_t=\exp(-e_t)$, so that smaller residuals yield larger scores.
The absolute compatibility score does not indicate whether task-relevant state evolution is improving over time.
We therefore define the dense reward from its temporal change:
\begin{equation}
r_t^{\mathrm{TD}}
=
\phi_t
-
\phi_{t-\Delta t}
=
\exp(-e_t)
-
\exp(-e_{t-\Delta t}),
\label{eq:td_reward}
\end{equation}
where $\Delta t$ is the temporal differencing stride.
Positive $r_t^{\mathrm{TD}}$ indicates increasing compatibility with the learned task dynamics, providing a directional signal for policy optimization.

During policy learning, DiTaR provides this TD reward as a dense shaping signal together with the sparse environment reward and the RND exploration bonus:
\begin{equation}
R_t
=
\lambda_{\mathrm{D}}
\tilde r_t^{\mathrm{TD}}
+
r_t^{\mathrm{env}}
+
\lambda_{\mathrm{expl}}
r_t^{\mathrm{RND}},
\label{eq:rl_reward}
\end{equation}
where
$\tilde r_t^{\mathrm{TD}}
=
(r_t^{\mathrm{TD}}-\mu_D)/\sigma_D$
is the normalized DiTaR reward,
and $\lambda_{\mathrm{D}}$ and $\lambda_{\mathrm{expl}}$ control the contributions of the learned reward and RND exploration bonus, respectively.

\begin{table}[t]
\centering
\small
\setlength{\tabcolsep}{3.5pt}
\caption{
\textbf{Aggregate RL performance on MetaWorld and ManiSkill.} We report final success rate, task coverage, and normalized time-to-threshold. $\uparrow$ higher is better, $\downarrow$ lower is better.
}
\label{tab:aggregate}
\resizebox{0.7\linewidth}{!}{
\begin{tabular}{llcccc}
\toprule
Benchmark & Method & Final SR $\uparrow$ & SR$\ge$80 $\uparrow$ & Reach 0.8 $\uparrow$ & NormT$@$0.8 $\downarrow$ \\
\midrule
\multirow{5}{*}{MetaWorld}
& Env-only & 56.8 & 50.0 & 50.0 & 0.68 \\
& VIPER & 39.9 & 33.3 & 41.7 & 0.73 \\
& Diff. Reward & 13.1 & 8.3 & 8.3 & 0.92 \\
& ReWiND & 65.2 & 33.3 & 41.7 & 0.61 \\
& \textbf{DiTaR} & \textbf{94.3} & \textbf{91.7} & \textbf{91.7} & \textbf{0.24} \\
\midrule
\multirow{5}{*}{ManiSkill}
& Env-only & 74.6 & 50.0 & 75.0 & 0.58 \\
& VIPER & 40.9 & 25.0 & 50.0 & 0.66 \\
& Diff. Reward & 30.8 & 0.0 & 25.0 & 0.88 \\
& ReWiND & 55.0 & 0.0 & 25.0 & 0.97 \\
& \textbf{DiTaR} & \textbf{87.5} & \textbf{100.0} & \textbf{100.0} & \textbf{0.43} \\
\bottomrule
\end{tabular}
}

\end{table}

\section{Experiments}

\subsection{Experimental Setup}

\paragraph{Benchmarks and Baselines.}
We evaluate DiTaR on 16 visual manipulation tasks from two standard simulation benchmarks: MetaWorld-v3~\citep{yu2020metaworld} and ManiSkill~\citep{gu2023maniskill2}, including 12 MetaWorld tasks and 4 ManiSkill tasks. We compare against four baselines: \textbf{Env-only}, which removes the learned reward term; \textbf{VIPER}~\citep{escontrela2023viper}, which uses conditional video-prediction likelihood as reward; \textbf{Diffusion Reward}~\citep{huang2024diffusionreward}, which derives rewards from a history-conditioned video diffusion model; and \textbf{ReWiND}~\citep{zhang2025rewind}, which learns a language-conditioned progress estimator using temporally transformed expert videos for supervision. We report task success rate as the primary evaluation metric.

\paragraph{Implementation Details.}
All methods use the same DrQ-v2~\citep{yaratsmastering} RL pipeline with $64\times64$ RGB observations, random-shift augmentation, and the same exploration setup, without privileged simulator states. The benchmark reward is retained for all methods, while learned rewards are added as dense shaping signals. VIPER, Diffusion Reward, and ReWiND are trained on expert videos from the corresponding MetaWorld or ManiSkill tasks. 
DiTaR instead pretrains a single task-relevant encoder and history- \& language-conditioned dynamics model on heterogeneous action-free videos, primarily from AgiBotWorld~\citep{bu2025agibot_iros}, DROID~\citep{khazatsky2024droid}, and Bridge~\citep{walke2023bridgedata}, with a small fraction from MetaWorld~\citep{mcleanmeta} and ManiSkill~\citep{gu2023maniskill2}. More details are in Appendices~\ref{sec:training_data} and~\ref{sec:suppl_policy_evaluation}.

\subsection{Main RL Performance}

\begin{figure*}[t]
    \centering
    \includegraphics[width=0.9\textwidth]{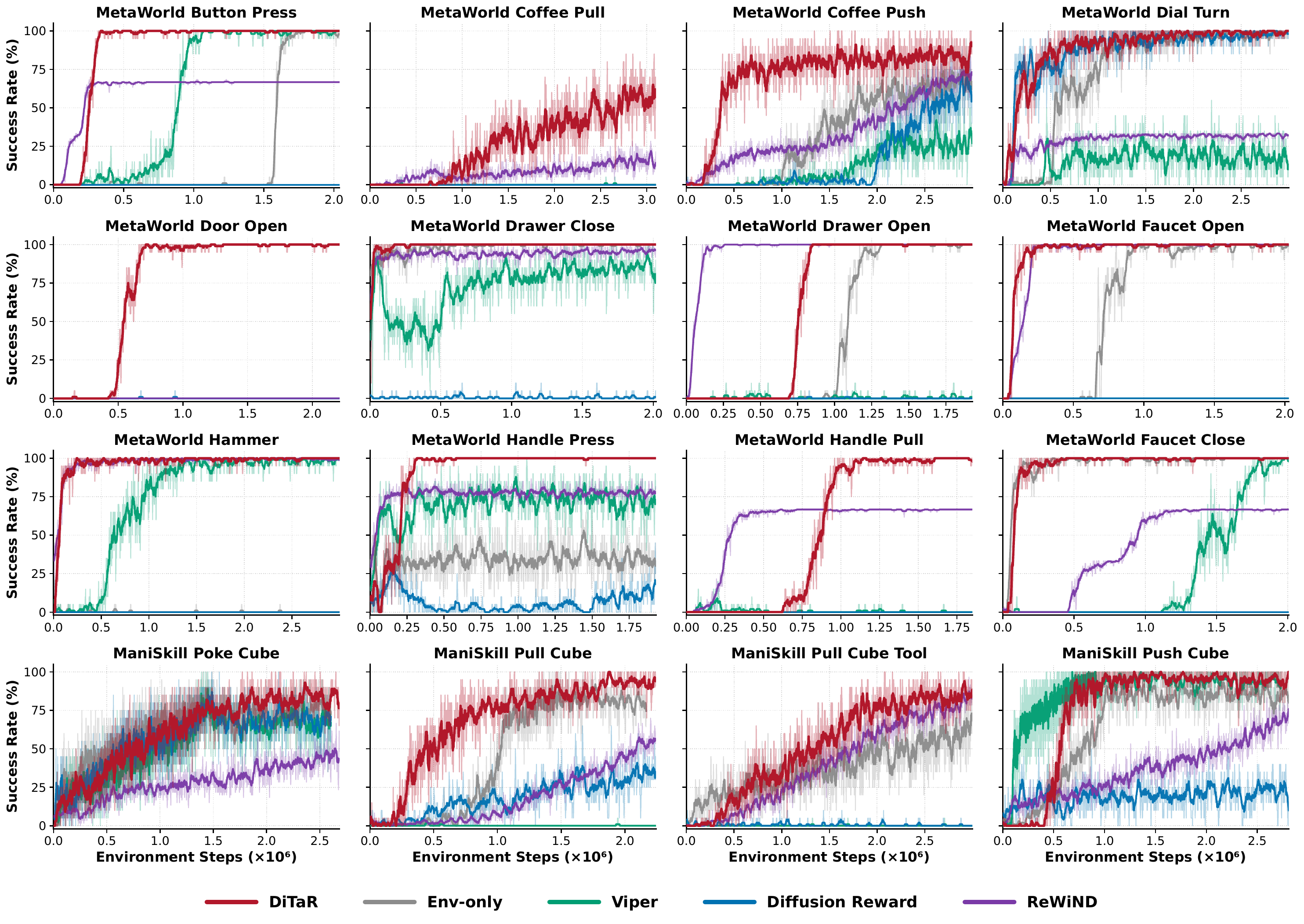}
    \caption{
    \textbf{Main RL performance.}
    DiTaR achieves strong and consistent policy learning across 12 MetaWorld and 4 ManiSkill tasks, outperforming Env-only, VIPER, Diffusion Reward, and ReWiND in aggregate performance. Panels use task-specific x-axis ranges for readability.
    }
    \label{fig:main results}

\end{figure*}

\begin{figure*}[t]
    \centering
    \includegraphics[width=\textwidth]{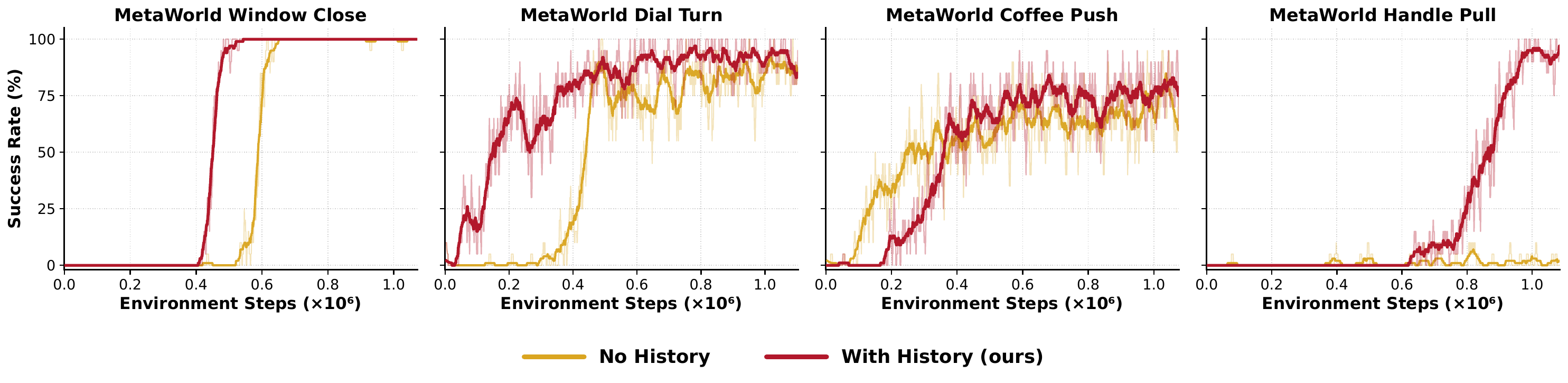}
    \caption{
    \textbf{Ablation on history conditioning.}
    Removing visual history from the conditional dynamics model reduces learning efficiency and final performance across the evaluated MetaWorld tasks.
    }
    \label{fig:ablation_history}

\end{figure*}

Figure~\ref{fig:main results} shows the per-task learning curves, and Table~\ref{tab:aggregate} summarizes the aggregate performance. Final SR denotes the success rate at the end of training. SR$\geq$80 is the percentage of tasks whose final success rate is at least 80\%, while Reach 0.8 is the percentage of tasks that reach 0.8 success at least once during training. NormT@0.8 is the normalized first hitting time to this threshold, assigning the full training budget when the threshold is not reached.

On MetaWorld, DiTaR achieves a final success rate of 94.3\%, substantially outperforming the strongest learned-reward baseline, ReWiND (65.2\%). It reaches 0.8 success on 91.7\% of tasks and achieves a NormT@0.8 of 0.24, compared with 0.61 for ReWiND. On ManiSkill, DiTaR achieves 87.5\% final success, exceeding Env-only (74.6\%) and ReWiND (55.0\%), while attaining 100\% Reach 0.8. Its NormT$@$0.8 of 0.43 is also lower than Env-only (0.58) and ReWiND (0.97). These results show that DiTaR improves both final policy performance and learning efficiency.
The per-task curves in Figure~\ref{fig:main results} show that DiTaR's aggregate gains come from more consistent learning across the task suite. ReWiND and VIPER perform well on some tasks but degrade substantially on others, whereas DiTaR reaches high success across a broader range of environments with fewer stalled or failed runs. This trend is consistent with the aggregate results in Table~\ref{tab:aggregate}.

\subsection{Ablation Studies}

\begin{figure*}[t]
    \centering
    \includegraphics[width=\textwidth]{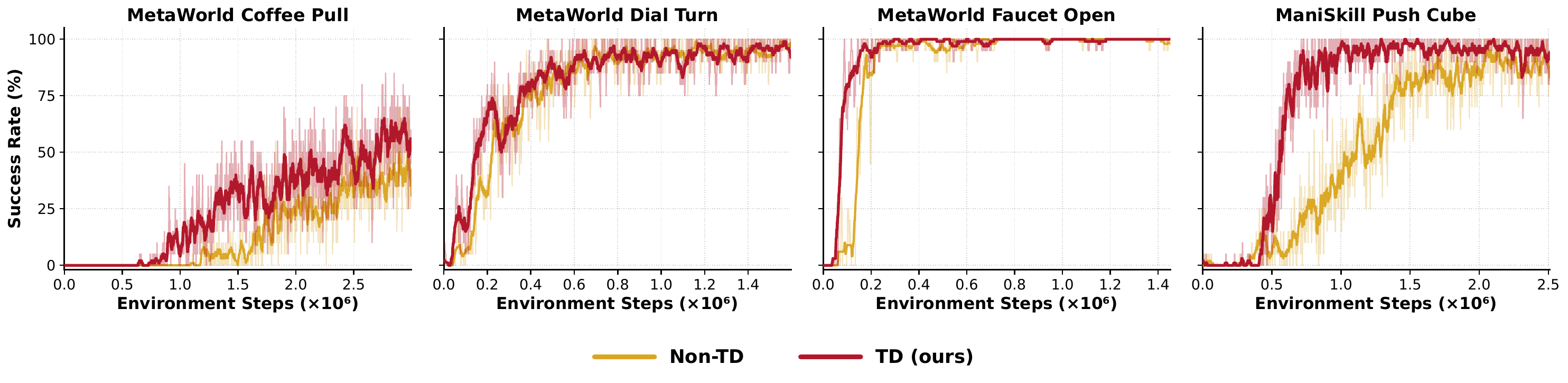}
    \caption{
    \textbf{Ablation on temporal-difference reward.}
    Temporal differencing improves policy learning over directly using the absolute compatibility score $\phi_t$ across MetaWorld and ManiSkill tasks.
    }
    \label{fig:ablation_td}

\end{figure*}

We ablate two designs in DiTaR: conditioning task-relevant dynamics on recent visual history and deriving directional rewards from temporal changes in dynamics compatibility.

\paragraph{History Conditioning.}
Figure~\ref{fig:ablation_history} compares DiTaR with a variant that removes visual history from the conditional dynamics model. History conditioning substantially accelerates learning on Window Close and Dial Turn and improves final performance on Coffee Push. The difference is particularly pronounced on Handle Pull, where the no-history variant makes little progress. These results show that history context is important for evaluating task-relevant transitions under the learned dynamics and provides a more effective reward signal for policy learning.

\paragraph{Temporal-Difference Reward.}
Figure~\ref{fig:ablation_td} compares DiTaR with a non-TD variant that directly uses the absolute compatibility score $\phi_t$ as reward. Temporal differencing improves learning efficiency on both MetaWorld and ManiSkill, with clear gains on Coffee Pull, Faucet Open, and ManiSkill Push Cube, while both variants achieve similar performance on Dial Turn. The compatibility score measures how well the current transition agrees with the learned task dynamics, but does not indicate whether this agreement is increasing over time. Using its temporal change therefore provides directional feedback that is more effective for policy optimization.
Appendix~\ref{sec:suppl_ablation_latent} analyzes the task-relevant latent via ablation (94.3\% vs. 67.2\% final SR) and latent grouping.

\subsection{Reward Analysis}

\begin{figure*}[t]
    \centering
    \includegraphics[width=\textwidth]{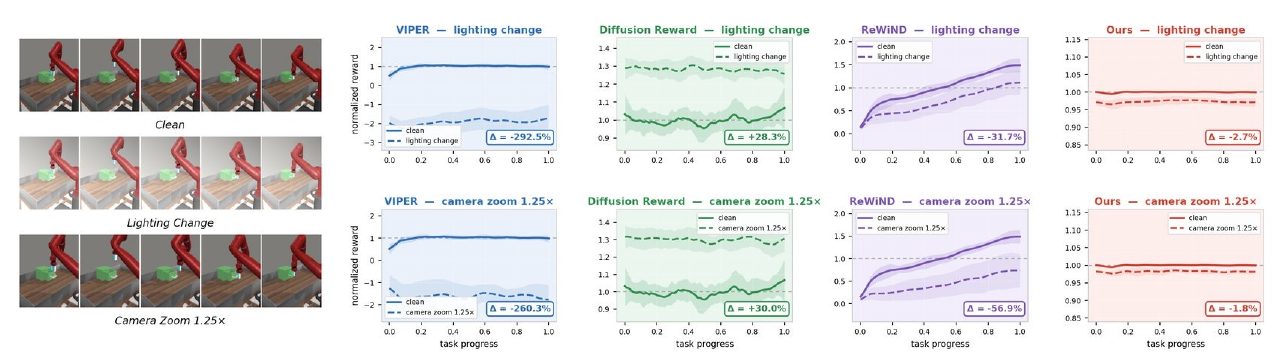}

    \caption{\textbf{Reward robustness to visual perturbations.} Per-frame reward trajectories on MetaWorld expert rollouts under lighting changes and $1.25\times$ camera zoom. $\Delta$ denotes the relative mean reward change from the clean rollout. DiTaR shows substantially smaller reward shifts than VIPER, Diffusion Reward, and ReWiND under both perturbations.}

    \label{fig:reward_robustness}
\end{figure*}

We evaluate two complementary properties of the DiTaR reward: sensitivity to task-relevant temporal structure and robustness to task-irrelevant visual variation. Table~\ref{tab:reward_analysis} reports reward changes under temporal and visual perturbations on MetaWorld and real-world expert videos. Figure~\ref{fig:reward_robustness} further shows representative reward trajectories under visual perturbations.

\begin{table}[t]
    \centering
    \caption{\textbf{Reward sensitivity and robustness.} For temporal perturbations, $\Delta$ denotes the signed reward change from clean trajectories, with more negative values indicating stronger temporal sensitivity. For visual perturbations, $|\Delta|$ denotes the reward-change magnitude, with smaller values indicating greater robustness. Diff. R.: Diffusion Reward.}
    \label{tab:reward_analysis}
    \vspace{3pt}
    \small
    \renewcommand{\arraystretch}{1.15}
    \resizebox{0.8\linewidth}{!}{%
    \begin{tabular}{lrrrrrrrr}
        \toprule
        & \multicolumn{4}{c}{\textbf{MetaWorld}}
        & \multicolumn{4}{c}{\textbf{Real-world}} \\
        \cmidrule(lr){2-5}\cmidrule(l){6-9}
        Perturbation
        & \textbf{DiTaR} & Diff. R. & VIPER & ReWiND
        & \textbf{DiTaR} & Diff. R. & VIPER & ReWiND \\
        \midrule
        \multicolumn{9}{l}{\textit{(a) Temporal sensitivity: $\Delta$ (\%) $\downarrow$}} \\
        \addlinespace[2pt]
        Locally shuffled
        & $\mathbf{-22.2}$ & $-0.6$ & $-10.2$ & $+0.1$
        & $-3.2$ & $+0.1$ & $\mathbf{-10.0}$ & $+0.2$ \\
        Temporal jitter $\pm2$
        & $\mathbf{-19.1}$ & $-0.3$ & $-4.7$ & $+1.5$
        & $-4.2$ & $-0.2$ & $\mathbf{-5.5}$ & $+0.6$ \\
        Shuffled chunks
        & $\mathbf{-28.9}$ & $+0.5$ & $-9.3$ & $-15.9$
        & $-7.0$ & $-1.6$ & $\mathbf{-11.8}$ & $+19.1$ \\
        \midrule
        \multicolumn{9}{l}{\textit{(b) Visual robustness: $|\Delta|$ (\%) $\downarrow$}} \\
        \addlinespace[2pt]
        Camera pose shift
        & $\mathbf{24.3}$ & $42.8$ & $44.6$ & $59.5$
        & $\mathbf{4.2}$ & $32.3$ & $10.2$ & $60.8$ \\
        Camera yaw $8^\circ$
        & $\mathbf{28.3}$ & $44.7$ & $40.9$ & $55.0$
        & $\mathbf{4.3}$ & $34.7$ & $16.1$ & $30.0$ \\
        Lighting (heavy)
        & $\mathbf{19.4}$ & $39.3$ & $49.7$ & $30.1$
        & $\mathbf{1.6}$ & $21.0$ & $29.5$ & $10.2$ \\
        \bottomrule
    \end{tabular}%
    }

\end{table}

\paragraph{Sensitivity to Temporal Structure.}
We perturb expert videos using local shuffling, temporal jitter, and shuffled chunks. These perturbations preserve the visual content of individual frames while disrupting the temporal order of task execution. On MetaWorld, DiTaR shows the largest reward decrease under all three perturbations, indicating strong sensitivity to changes in task execution. On real-world videos, DiTaR also consistently assigns lower rewards to temporally corrupted trajectories, although VIPER shows larger decreases. Overall, DiTaR remains sensitive to disruptions in temporal task structure across both settings.

\paragraph{Robustness to Visual Variation.}
We next perturb camera viewpoint and lighting while keeping the underlying task execution unchanged. In this setting, a desirable reward should remain close to its value on the clean trajectory. DiTaR shows the smallest reward shift across all evaluated visual perturbations on both MetaWorld and real-world videos, as shown in Table~\ref{tab:reward_analysis}. Figure~\ref{fig:reward_robustness} shows the same behavior for representative lighting and camera-zoom perturbations: the competing rewards change substantially, whereas DiTaR remains comparatively stable. Together, these results show that DiTaR responds to changes in temporal task structure while being substantially less affected by task-irrelevant visual variation.
Additional analyses of temporal corruptions, corruption localization, viewpoint robustness, and reward evolution during policy training are provided in Appendix~\ref{sec:reward_analysis}.

\subsection{Real-World Offline RL}
\label{sec:realworld_offline}

\begin{figure*}[!tbp]
    \centering
    \includegraphics[width=0.6\textwidth]{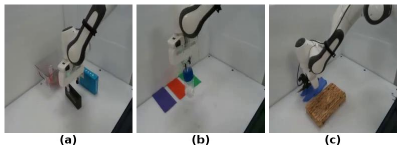}
    \caption{\textbf{Real-world task suite.} Three tabletop manipulation tasks used for offline RL evaluation: (a) object-to-basket placement, (b) bottle-to-plate placement, and (c) knife-to-glove placement.}
    \label{fig:real_world_demo}
\end{figure*}

We further evaluate DiTaR on real-world offline RL using 150 demonstrations collected with a Franka Research 3 robot across three tabletop manipulation tasks (Figure~\ref{fig:real_world_demo}). Each demonstration contains synchronized third-person and wrist RGB observations.
For each reward model, we relabel the same offline dataset and train a policy using Implicit Q-Learning (IQL)~\citep{kostrikov2022offline}. Thus, VIPER, Diffusion Reward, ReWiND, and DiTaR differ only in the reward assigned to the fixed dataset, while sharing the same demonstrations and IQL training pipeline. As shown in Table~\ref{tab:real_offline_rl}, DiTaR achieves an overall success rate of 83.3\%, compared with 26.7\% for Diffusion Reward, 16.7\% for VIPER, and 43.3\% for ReWiND.

\begin{table}[!tbp]
\centering
\small
\caption{\textbf{Real-world offline RL results.} All methods relabel the same offline demonstrations and use the same IQL training pipeline. We conduct 10 evaluation trials per task.}
\label{tab:real_offline_rl}
\vspace{2mm}
\begin{tabular}{lcccc}
\toprule
Metric & VIPER & Diffusion Reward & ReWiND & \textbf{DiTaR} \\
\midrule
Success Rate & 16.7\% & 26.7\% & 43.3\% & \textbf{83.3\%} \\
\bottomrule
\end{tabular}

\end{table}

\section{Conclusion and Limitations}

DiTaR learns dense rewards by modeling task-relevant latent dynamics and temporal changes in dynamics compatibility. It improves policy learning on MetaWorld-v3 and ManiSkill and transfers to real-world offline RL through reward relabeling. Reward analyses show sensitivity to task-relevant temporal structure and robustness to task-irrelevant visual variation. However, DiTaR still relies on sparse environment rewards during RL, leaving fully video-based reward learning for future work. Its pretraining data are also dominated by limited-view robotic manipulation videos; broader multi-view and egocentric data may further improve generalization across diverse manipulation tasks and environments.

\clearpage

\bibliographystyle{assets/plainnat}
\bibliography{references}

@inproceedings{escontrela2023viper,
  title     = {Video Prediction Models as Rewards for Reinforcement Learning},
  author    = {Escontrela, Alejandro and Adeniji, Ademi and Yan, Wilson and Jain, Ajay and Peng, Xue Bin and Goldberg, Ken and Lee, Youngwoon and Hafner, Danijar and Abbeel, Pieter},
  booktitle = {Advances in Neural Information Processing Systems},
  year      = {2023}
}

@inproceedings{huang2024diffusionreward,
  title     = {Diffusion Reward: Learning Rewards via Conditional Video Diffusion},
  author    = {Huang, Tao and Jiang, Guangqi and Ze, Yanjie and Xu, Huazhe},
  booktitle = {European Conference on Computer Vision},
  year      = {2024}
}

@inproceedings{zhang2025rewind,
  title     = {{ReWiND}: Language-Guided Rewards Teach Robot Policies without New Demonstrations},
  author    = {Zhang, Jiahui and Luo, Yusen and Anwar, Abrar and
               Sontakke, Sumedh Anand and Lim, Joseph J and Thomason, Jesse and
               Biyik, Erdem and Zhang, Jesse},
  booktitle = {Proceedings of The 9th Conference on Robot Learning},
  pages     = {460--488},
  year      = {2025},
  volume    = {305},
  series    = {Proceedings of Machine Learning Research},
  publisher = {PMLR}
}

@inproceedings{tan2025robodopamine,
  author    = {Tan, Huajie and Chen, Sixiang and Xu, Yijie and Wang, Zixiao and
               Chi, Cheng and Ji, Yuheng and Lyu, Yaoxu and Zhao, Zhongxia and
               Chen, Xiansheng and Co, Peterson and Xie, Shaoxuan and Yao, Guocai and
               Wang, Pengwei and Wang, Zhongyuan and Zhang, Shanghang},
  title     = {General Process Reward Modeling for Robotic Reinforcement Learning},
  booktitle = {Proceedings of the IEEE/CVF Conference on Computer Vision and Pattern Recognition},
  pages     = {22412--22422},
  year      = {2026}
}

@inproceedings{wang2025genreward,
  author    = {Wang, Qi and Wu, Mian and Zhang, Yuyang and Yuan, Mingqi and
               Zhang, Wenyao and You, Haoxiang and Wang, Yunbo and Jin, Xin and
               Yang, Xiaokang and Zeng, Wenjun},
  title     = {Goal-Driven Reward by Video Diffusion Models for Reinforcement Learning},
  booktitle = {Proceedings of the IEEE/CVF Conference on Computer Vision and Pattern Recognition},
  year      = {2026}
}

@article{bu2025univla,
  title   = {Learning to Act Anywhere with Task-centric Latent Actions},
  author  = {Bu, Qingwen and Yang, Yanting and Cai, Jisong and Gao, Shenyuan and Ren, Guanghui and Yao, Maoqing and Luo, Ping and Li, Hongyang},
  journal = {arXiv preprint arXiv:2505.06111},
  year    = {2025}
}

@inproceedings{yu2020metaworld,
  title     = {{Meta-World}: A Benchmark and Evaluation for Multi-Task and Meta Reinforcement Learning},
  author    = {Yu, Tianhe and Quillen, Deirdre and He, Zhanpeng and Julian, Ryan and Hausman, Karol and Finn, Chelsea and Levine, Sergey},
  booktitle = {Proceedings of the Conference on Robot Learning},
  pages     = {1094--1100},
  year      = {2020}
}

@inproceedings{gu2023maniskill2,
  title     = {{ManiSkill2}: A Unified Benchmark for Generalizable Manipulation Skills},
  author    = {Gu, Jiayuan and Xiang, Fanbo and Li, Xuanlin and Ling, Zhan and Liu, Xiqiang and Mu, Tongzhou and Tang, Yihe and Tao, Stone and Wei, Xinyue and Yao, Yunchao and Yuan, Xiaodi and Xie, Pengwei and Huang, Zhiao and Chen, Rui and Su, Hao},
  booktitle = {International Conference on Learning Representations},
  year      = {2023}
}

@article{lee2026roboreward,
  title   = {RoboReward: General-Purpose Vision-Language Reward Models for Robotics},
  author  = {Lee, Tony and Wagenmaker, Andrew and Pertsch, Karl and Liang, Percy and Levine, Sergey and Finn, Chelsea},
  journal = {arXiv preprint arXiv:2601.00675},
  year    = {2026}
}

@article{zhai2025vlac,
  title   = {A Vision-Language-Action-Critic Model for Robotic Real-World Reinforcement Learning},
  author  = {Zhai, Shaopeng and Zhang, Qi and Zhang, Tianyi and Huang, Fuxian and Zhang, Haoran and Zhou, Ming and Zhang, Shengzhe and Liu, Litao and Lin, Sixu and Pang, Jiangmiao},
  journal = {arXiv preprint arXiv:2509.15937},
  year    = {2025}
}

@inproceedings{chen2025sarm,
  title     = {{SARM}: Stage-Aware Reward Modeling for Long Horizon Robot Manipulation},
  author    = {Chen, Qianzhong and Yu, Justin and Schwager, Mac and Abbeel, Pieter and
               Shentu, Yide and Wu, Philipp},
  booktitle = {International Conference on Learning Representations},
  year      = {2026}
}

@inproceedings{zhou2025dinowm,
  title     = {{DINO-WM}: World Models on Pre-trained Visual Features Enable Zero-shot Planning},
  author    = {Zhou, Gaoyue and Pan, Hengkai and LeCun, Yann and Pinto, Lerrel},
  booktitle = {Proceedings of the 42nd International Conference on Machine Learning},
  series    = {Proceedings of Machine Learning Research},
  volume    = {267},
  pages     = {79115--79135},
  year      = {2025}
}

@inproceedings{rocamonde2024vlmrm,
  title     = {Vision-Language Models are Zero-Shot Reward Models for Reinforcement Learning},
  author    = {Rocamonde, Juan and Montesinos, Victoriano and Nava, Elvis and Perez, Ethan and Lindner, David},
  booktitle = {International Conference on Learning Representations},
  year      = {2024}
}

@article{sontakke2023roboclip,
  title   = {{RoboCLIP}: One Demonstration is Enough to Learn Robot Policies},
  author  = {Sontakke, Sumedh and Zhang, Jesse and Arnold, S{\'e}b and Pertsch, Karl and B{\i}y{\i}k, Erdem and Sadigh, Dorsa and Finn, Chelsea and Itti, Laurent},
  journal = {Advances in Neural Information Processing Systems},
  volume  = {36},
  pages   = {55681--55693},
  year    = {2023}
}

@inproceedings{ma2023liv,
  title        = {{LIV}: Language-Image Representations and Rewards for Robotic Control},
  author       = {Ma, Yecheng Jason and Kumar, Vikash and Zhang, Amy and Bastani, Osbert and Jayaraman, Dinesh},
  booktitle    = {International Conference on Machine Learning},
  pages        = {23301--23320},
  year         = {2023},
  organization = {PMLR}
}

@article{alakuijala2024video,
  title   = {Video-Language Critic: Transferable Reward Functions for Language-Conditioned Robotics},
  author  = {Alakuijala, Minttu and McLean, Reginald and Woungang, Isaac and Farsad, Nariman and Kaski, Samuel and Marttinen, Pekka and Yuan, Kai},
  journal = {arXiv preprint arXiv:2405.19988},
  year    = {2024}
}

@article{luo2024text,
  title   = {Text-Aware Diffusion for Policy Learning},
  author  = {Luo, Calvin and He, Mandy and Zeng, Zilai and Sun, Chen},
  journal = {Advances in Neural Information Processing Systems},
  volume  = {37},
  pages   = {46226--46253},
  year    = {2024}
}

@inproceedings{ma2023vip,
  title     = {{VIP}: Towards Universal Visual Reward and Representation via Value-Implicit Pre-Training},
  author    = {Ma, Yecheng Jason and Sodhani, Shagun and Jayaraman, Dinesh and Bastani, Osbert and Kumar, Vikash and Zhang, Amy},
  booktitle = {The Eleventh International Conference on Learning Representations},
  year      = {2023}
}

@inproceedings{nair2023r3m,
  title        = {{R3M}: A Universal Visual Representation for Robot Manipulation},
  author       = {Nair, Suraj and Rajeswaran, Aravind and Kumar, Vikash and Finn, Chelsea and Gupta, Abhinav},
  booktitle    = {Conference on Robot Learning},
  pages        = {892--909},
  year         = {2023},
  organization = {PMLR}
}

@inproceedings{karamcheti2023language,
  author    = {Siddharth Karamcheti and Suraj Nair and Annie S Chen and Thomas Kollar and
               Chelsea Finn and Dorsa Sadigh and Percy Liang},
  title     = {{Language-Driven Representation Learning for Robotics}},
  booktitle = {Proceedings of Robotics: Science and Systems},
  year      = {2023},
  address   = {Daegu, Republic of Korea},
  month     = {July},
  doi       = {10.15607/RSS.2023.XIX.032}
}

@article{assran2025v,
  title   = {{V-JEPA 2}: Self-Supervised Video Models Enable Understanding, Prediction and Planning},
  author  = {Assran, Mido and Bardes, Adrien and Fan, David and Garrido, Quentin and
             Howes, Russell and Komeili, Mojtaba and others},
  journal = {arXiv preprint arXiv:2506.09985},
  year    = {2025}
}

@article{mur2026v,
  title   = {{V-JEPA} 2.1: Unlocking Dense Features in Video Self-Supervised Learning},
  author  = {Mur-Labadia, Lorenzo and Muckley, Matthew and Bar, Amir and Assran, Mido and Sinha, Koustuv and Rabbat, Mike and LeCun, Yann and Ballas, Nicolas and Bardes, Adrien},
  journal = {arXiv preprint arXiv:2603.14482},
  year    = {2026}
}

@article{maes2026leworldmodel,
  title   = {{LeWorldModel}: Stable End-to-End Joint-Embedding Predictive Architecture from Pixels},
  author  = {Maes, Lucas and Le Lidec, Quentin and Scieur, Damien and LeCun, Yann and Balestriero, Randall},
  journal = {arXiv preprint arXiv:2603.19312},
  year    = {2026}
}

@inproceedings{lyu2026lda,
  author    = {Jiangran Lyu and Kai Liu and Xuheng Zhang and Haoran Liao and Yusen Feng and
               Wenxuan Zhu and Tingrui Shen and Jiayi Chen and Jiazhao Zhang and Yifei Dong and
               Cui Wenbo and Senmao Qi and Shuo Wang and Yixin Zheng and Mi Yan and Xuesong Shi and
               Haoran Li and Dongbin Zhao and Ming-Yu Liu and Zhizheng Zhang and Li Yi and
               Yizhou Wang and He Wang},
  title     = {{LDA-1B: Scaling Latent Dynamics Action Model via Universal Embodied Data Ingestion}},
  booktitle = {Proceedings of Robotics: Science and Systems},
  year      = {2026},
  address   = {Sydney, Australia},
  month     = {July},
  doi       = {10.15607/RSS.2026.XXII.210}
}

@article{zhang2026clap,
  title   = {{CLAP}: Contrastive Latent Action Pretraining for Learning Vision-Language-Action Models from Human Videos},
  author  = {Zhang, Chubin and Wang, Jianan and Gao, Zifeng and Su, Yue and Dai, Tianru and Zhou, Cai and Lu, Jiwen and Tang, Yansong},
  journal = {arXiv preprint arXiv:2601.04061},
  year    = {2026}
}

@inproceedings{jia2026guidedvlaspecifyingtaskrelevantfactors,
  author    = {Xiaosong Jia and Bowen Yang and Zuhao Ge and Xian Nie and Yuchen Zhou and
               Cunxin Fan and Yufeng Li and Yilin Chai and Chao Jing and Zijian Liang and
               Qingwen Bu and Haidong Cao and Chao Wu and Qifeng Li and Zhenjie Yang and
               Chenhe Zhang and Hongyang Li and Zuxuan Wu and Junchi Yan and Yu-Gang Jiang},
  title     = {{GuidedVLA: Specifying Task-Relevant Factors via Plug-and-Play Action Attention Specialization}},
  booktitle = {Proceedings of Robotics: Science and Systems},
  year      = {2026},
  address   = {Sydney, Australia},
  month     = {July},
  doi       = {10.15607/RSS.2026.XXII.084}
}

@article{raffel2020exploring,
  title={Exploring the limits of transfer learning with a unified text-to-text transformer},
  author={Raffel, Colin and Shazeer, Noam and Roberts, Adam and Lee, Katherine and Narang, Sharan and Matena, Michael and Zhou, Yanqi and Li, Wei and Liu, Peter J},
  journal={Journal of machine learning research},
  volume={21},
  number={140},
  pages={1--67},
  year={2020}
}

@inproceedings{yaratsmastering,
  title={Mastering Visual Continuous Control: Improved Data-Augmented Reinforcement Learning},
  author={Yarats, Denis and Fergus, Rob and Lazaric, Alessandro and Pinto, Lerrel},
  booktitle={International Conference on Learning Representations},
  year={2022}
}

@inproceedings{bu2025agibot_iros,
  title={Agibot world colosseo: A large-scale manipulation platform for scalable and intelligent embodied systems},
  author={Bu, Qingwen and Cai, Jisong and Chen, Li and Cui, Xiuqi and Ding, Yan and Feng, Siyuan and He, Xindong and Huang, Xu and others},
  booktitle={2025 IEEE/RSJ International Conference on Intelligent Robots and Systems (IROS)},
  year={2025},
  organization={IEEE}
}

@inproceedings{khazatsky2024droid,
  title={DROID: A large-scale in-the-wild robot manipulation dataset},
  author={Khazatsky, Alexander and Pertsch, Karl and Nair, Suraj and Balakrishna, Ashwin and Dasari, Sudeep and Karamcheti, Siddharth and Nasiriany, Soroush and Srirama, Mohan Kumar and Chen, Lawrence Yunliang and Ellis, Kirsty and others},
  booktitle={Robotics: Science and Systems},
  year={2024}
}

@inproceedings{walke2023bridgedata,
  title={Bridgedata v2: A dataset for robot learning at scale},
  author={Walke, Homer Rich and Black, Kevin and Zhao, Tony Z and Vuong, Quan and Zheng, Chongyi and Hansen-Estruch, Philippe and He, Andre Wang and Myers, Vivek and Kim, Moo Jin and Du, Max and others},
  booktitle={Conference on Robot Learning},
  pages={1723--1736},
  year={2023},
  organization={PMLR}
}

@inproceedings{mcleanmeta,
  title     = {{Meta-World+}: An Improved, Standardized, RL Benchmark},
  author    = {McLean, Reginald and Chatzaroulas, Evangelos and McCutcheon, Luc and
               R{\"o}der, Frank and Yu, Tianhe and He, Zhanpeng and Zentner, K. R. and
               Julian, Ryan and Terry, J K and Woungang, Isaac and Farsad, Nariman and
               Castro, Pablo Samuel},
  booktitle = {Advances in Neural Information Processing Systems},
  volume    = {38},
  year      = {2025}
}

@article{wan2025wan,
  title={Wan: Open and advanced large-scale video generative models},
  author={Wan, Team and Wang, Ang and Ai, Baole and Wen, Bin and Mao, Chaojie and Xie, Chen-Wei and Chen, Di and Yu, Feiwu and Zhao, Haiming and Yang, Jianxiao and others},
  journal={arXiv preprint arXiv:2503.20314},
  year={2025}
}

@article{aytar2018playing,
  title={Playing hard exploration games by watching youtube},
  author={Aytar, Yusuf and Pfaff, Tobias and Budden, David and Paine, Thomas and Wang, Ziyu and De Freitas, Nando},
  journal={Advances in neural information processing systems},
  volume={31},
  year={2018}
}

@inproceedings{sermanet2018time,
  title={Time-contrastive networks: Self-supervised learning from video},
  author={Sermanet, Pierre and Lynch, Corey and Chebotar, Yevgen and Hsu, Jasmine and Jang, Eric and Schaal, Stefan and Levine, Sergey},
  booktitle={2018 IEEE international conference on robotics and automation (ICRA)},
  pages={1134--1141},
  year={2018},
  organization={IEEE}
}

@inproceedings{dwibedi2019temporal,
  title={Temporal cycle-consistency learning},
  author={Dwibedi, Debidatta and Aytar, Yusuf and Tompson, Jonathan and Sermanet, Pierre and Zisserman, Andrew},
  booktitle={Proceedings of the IEEE/CVF conference on computer vision and pattern recognition},
  pages={1801--1810},
  year={2019}
}

@inproceedings{chen2021learning,
  author    = {Annie S. Chen and Suraj Nair and Chelsea Finn},
  title     = {{Learning Generalizable Robotic Reward Functions from ``In-The-Wild'' Human Videos}},
  booktitle = {Proceedings of Robotics: Science and Systems},
  year      = {2021},
  address   = {Virtual},
  month     = {July},
  doi       = {10.15607/RSS.2021.XVII.012}
}

@inproceedings{alakuijala2023learning,
  title={Learning reward functions for robotic manipulation by observing humans},
  author={Alakuijala, Minttu and Dulac-Arnold, Gabriel and Mairal, Julien and Ponce, Jean and Schmid, Cordelia},
  booktitle={2023 IEEE International Conference on Robotics and Automation (ICRA)},
  pages={5006--5012},
  year={2023},
  organization={IEEE}
}

@inproceedings{yang2024rank2reward,
  title={Rank2reward: Learning shaped reward functions from passive video},
  author={Yang, Daniel and Tjia, Davin and Berg, Jacob and Damen, Dima and Agrawal, Pulkit and Gupta, Abhishek},
  booktitle={2024 IEEE International Conference on Robotics and Automation (ICRA)},
  pages={2806--2813},
  year={2024},
  organization={IEEE}
}

@inproceedings{yang2024adapt2reward,
  title={Adapt2reward: Adapting video-language models to generalizable robotic rewards via failure prompts},
  author={Yang, Yanting and Chen, Minghao and Qiu, Qibo and Wu, Jiahao and Wang, Wenxiao and Lin, Binbin and Guan, Ziyu and He, Xiaofei},
  booktitle={European Conference on Computer Vision},
  pages={163--180},
  year={2024},
  organization={Springer}
}

@inproceedings{biza2025robot,
  title={On-Robot Reinforcement Learning with Goal-Contrastive Rewards},
  author={Biza, Ondrej and Weng, Thomas and Sun, Lingfeng and Schmeckpeper, Karl and Kelestemur, Tarik and Ma, Yecheng Jason and Platt, Robert and van de Meent, Jan-Willem and Wong, Lawson LS},
  booktitle={2025 IEEE International Conference on Robotics and Automation (ICRA)},
  pages={4797--4805},
  year={2025},
  organization={IEEE}
}

@inproceedings{ayalew2025progressor,
  title={Progressor: A perceptually guided reward estimator with self-supervised online refinement},
  author={Ayalew, Tewodros W and Zhang, Xiao and Wu, Kevin Yuanbo and Jiang, Tianchong and Maire, Michael and Walter, Matthew R},
  booktitle={Proceedings of the IEEE/CVF International Conference on Computer Vision},
  pages={10297--10306},
  year={2025}
}

@inproceedings{ma2025vision,
  title={Vision language models are in-context value learners},
  author={Ma, Yecheng Jason and Hejna, Joey and Fu, Chuyuan and Shah, Dhruv and Liang, Jacky and Xu, Zhuo and Kirmani, Sean and Xu, Peng and Driess, Danny and Xiao, Ted and others},
  booktitle={International Conference on Learning Representations},
  volume={2025},
  pages={33984--34009},
  year={2025}
}

@inproceedings{hung2025victor,
  title={VICtoR: Learning hierarchical vision-instruction correlation rewards for long-horizon manipulation},
  author={Hung, Kuo-Han and Lo, Pang-Chi and Yeh, Jia-Fong and Hsu, Han-Yuan and Chen, Yi-Ting and Hsu, Winston},
  booktitle={International Conference on Learning Representations},
  volume={2025},
  pages={17550--17577},
  year={2025}
}

@inproceedings{kim2025subtask,
  title={Subtask-aware visual reward learning from segmented demonstrations},
  author={Kim, Changyeon and Heo, Minho and Lee, Doohyun and Lee, Honglak and Shin, Jinwoo and Lim, Joseph and Lee, Kimin},
  booktitle={International Conference on Learning Representations},
  volume={2025},
  pages={15142--15165},
  year={2025}
}

@inproceedings{assran2023self,
  title={Self-supervised learning from images with a joint-embedding predictive architecture},
  author={Assran, Mahmoud and Duval, Quentin and Misra, Ishan and Bojanowski, Piotr and Vincent, Pascal and Rabbat, Michael and LeCun, Yann and Ballas, Nicolas},
  booktitle={Proceedings of the IEEE/CVF conference on computer vision and pattern recognition},
  pages={15619--15629},
  year={2023}
}

@article{bardesrevisiting,
  title={Revisiting Feature Prediction for Learning Visual Representations from Video},
  author={Bardes, Adrien and Garrido, Quentin and Ponce, Jean and Chen, Xinlei and Rabbat, Michael and LeCun, Yann and Assran, Mido and Ballas, Nicolas},
  journal={Transactions on Machine Learning Research},
  year={2024}
}

@article{oquab2024dinov2,
  title={{DINOv2}: Learning Robust Visual Features without Supervision},
  author={Oquab, Maxime and Darcet, Timoth{\'e}e and Moutakanni, Th{\'e}o and Vo, Huy and Szafraniec, Marc and Khalidov, Vasil and Fernandez, Pierre and Haziza, Daniel and Massa, Francisco and El-Nouby, Alaaeldin and others},
  journal={Transactions on Machine Learning Research},
  year={2024}
}

@article{hafner2023mastering,
  title={Mastering diverse domains through world models},
  author={Hafner, Danijar and Pasukonis, Jurgis and Ba, Jimmy and Lillicrap, Timothy},
  journal={arXiv preprint arXiv:2301.04104},
  year={2023}
}

@inproceedings{hansen2024td,
  title={Td-mpc2: Scalable, robust world models for continuous control},
  author={Hansen, Nick and Su, Hao and Wang, Xiaolong},
  booktitle={International Conference on Learning Representations},
  volume={2024},
  pages={47376--47405},
  year={2024}
}

@inproceedings{bruce2024genie,
  title={Genie: Generative interactive environments},
  author={Bruce, Jake and Dennis, Michael D and Edwards, Ashley and Parker-Holder, Jack and Shi, Yuge and Hughes, Edward and Lai, Matthew and Mavalankar, Aditi and Steigerwald, Richie and Apps, Chris and others},
  booktitle={Forty-first International Conference on Machine Learning},
  year={2024}
}

@article{du2023learning,
  title={Learning universal policies via text-guided video generation},
  author={Du, Yilun and Yang, Sherry and Dai, Bo and Dai, Hanjun and Nachum, Ofir and Tenenbaum, Josh and Schuurmans, Dale and Abbeel, Pieter},
  journal={Advances in neural information processing systems},
  volume={36},
  pages={9156--9172},
  year={2023}
}

@inproceedings{ye2025latent,
  title={Latent action pretraining from videos},
  author={Ye, Seonghyeon and Jang, Joel and Jeon, Byeongguk and Joo, Se June and Yang, Jianwei and Peng, Baolin and Mandlekar, Ajay and Tan, Reuben and Chao, Yu-Wei and Lin, Bill Yuchen and others},
  booktitle={International Conference on Learning Representations},
  volume={2025},
  pages={28213--28239},
  year={2025}
}

@inproceedings{luo2025grounding,
  title={Grounding video models to actions through goal conditioned exploration},
  author={Luo, Yunhao and Du, Yilun},
  booktitle={International Conference on Learning Representations},
  volume={2025},
  pages={92200--92232},
  year={2025}
}

@inproceedings{bharadhwaj2024track2act,
  title={Track2act: Predicting point tracks from internet videos enables generalizable robot manipulation},
  author={Bharadhwaj, Homanga and Mottaghi, Roozbeh and Gupta, Abhinav and Tulsiani, Shubham},
  booktitle={European Conference on Computer Vision},
  pages={306--324},
  year={2024},
  organization={Springer}
}

@inproceedings{liu2025timerewarder,
  title     = {{TimeRewarder}: Learning Dense Reward from Passive Videos via Frame-wise Temporal Distance},
  author    = {Liu, Yuyang and Wen, Chuan and Hu, Yihang and Jayaraman, Dinesh and Gao, Yang},
  booktitle = {International Conference on Machine Learning},
  year      = {2026}
}

@inproceedings{esser2024scaling,
  title     = {Scaling Rectified Flow Transformers for High-Resolution Image Synthesis},
  author    = {Esser, Patrick and Kulal, Sumith and Blattmann, Andreas and Entezari, Rahim
               and M{\"u}ller, Jonas and Saini, Harry and Levi, Yam and Lorenz, Dominik
               and Sauer, Axel and Boesel, Frederic and Podell, Dustin and Dockhorn, Tim
               and English, Zion and Rombach, Robin},
  booktitle = {Proceedings of the 41st International Conference on Machine Learning (ICML)},
  pages     = {12606--12633},
  year      = {2024},
  organization = {PMLR}
}

@inproceedings{kostrikov2022offline,
  title={Offline Reinforcement Learning with Implicit Q-Learning},
  author={Kostrikov, Ilya and Nair, Ashvin and Levine, Sergey},
  booktitle={International Conference on Learning Representations},
  year={2022}
}

\clearpage
\appendix
\section{Overview}
This supplementary material provides additional experimental analysis and implementation details for DiTaR. Sec.~\ref{sec:training_data} first describes the pretraining corpus, including the data sources, camera views, and temporal sampling used to train the task-relevant encoder and conditional dynamics model. Sec.~\ref{sec:reward_analysis} then extends the reward analysis from the main paper with temporal corruption tests, localization of corrupted segments, robustness to viewpoint changes, and reward evolution during policy training. Sec.~\ref{sec:suppl_ablation_latent} studies the task-relevant representation through policy-learning ablations and cross-dataset latent grouping. Sec.~\ref{sec:suppl_impl} gives the model and reward implementation details, including the latent encoders, conditional rectified-flow model, and reward computation. Sec.~\ref{sec:suppl_policy_evaluation} describes the online RL and real-world IQL settings, baseline implementations, and evaluation metrics. Finally, Sec.~\ref{sec:suppl_algorithms} gives pseudocode for representation learning, dynamics training, and online policy learning.

\section{Training Data Details}
\label{sec:training_data}

Table~\ref{tab:training_data} summarizes the action-free video data used for DiTaR pretraining. We train the task-relevant encoder and history- \& language-conditioned dynamics model on a mixture of real-world robot datasets and simulated manipulation environments. The pretraining corpus is dominated by real-world robot videos from AgiBotWorld~\citep{bu2025agibot_iros}, DROID~\citep{khazatsky2024droid}, and Bridge~\citep{walke2023bridgedata}, which account for 38.0\%, 38.0\%, and 19.0\% of the data, respectively. These datasets cover different robot embodiments, manipulation behaviors, and camera viewpoints, including head-mounted and exterior views. We additionally include MetaWorld~\citep{mcleanmeta} and ManiSkill~\citep{gu2023maniskill2}, each contributing 2.5\% of the pretraining data.

Across all sources, DiTaR uses RGB observations and paired language instructions without action labels or other embodiment-specific annotations. This allows the same representation and dynamics objectives to be applied across datasets with different robots and action spaces. Simulated videos account for a small fraction of pretraining; most videos come from real-world robot datasets.

The native frame rates differ across datasets, so we sample clips with dataset-specific frame strides to obtain comparable temporal coverage. We draw strides from $\{1,2,3,4\}$ for AgiBotWorld and DROID, $\{1,2,3\}$ for Bridge, $\{1,2\}$ for MetaWorld, and use a fixed stride of $1$ for ManiSkill. Sampling is biased toward the largest stride in each range to include longer-horizon temporal context while retaining shorter transitions.

\begin{table}[h]
    \centering
    \footnotesize
    \setlength{\tabcolsep}{5pt}
    \renewcommand{\arraystretch}{1.05}
    \caption{Video sources, proportions, camera views, and frame strides used for DiTaR pretraining.}
    \label{tab:training_data}
    \begin{tabular}{lccc}
        \toprule
        Dataset & Ratio (\%) & View & Frame stride \\
        \midrule
        AgiBotWorld~\citep{bu2025agibot_iros} & 38.0 & Head & $\{1,2,3,4\}$ \\
        DROID~\citep{khazatsky2024droid} & 38.0 & Exterior & $\{1,2,3,4\}$ \\
        Bridge~\citep{walke2023bridgedata} & 19.0 & Exterior & $\{1,2,3\}$ \\
        MetaWorld~\citep{mcleanmeta} & 2.5 & Left & $\{1,2\}$ \\
        ManiSkill~\citep{gu2023maniskill2} & 2.5 & Default & $1$ \\
        \bottomrule
    \end{tabular}
\end{table}

\section{Reward Analysis}
\label{sec:reward_analysis}

We examine how the DiTaR reward responds to task-relevant temporal changes and task-irrelevant visual variation.
Temporal perturbations disrupt the order or continuity of task execution, whereas visual perturbations alter appearance while preserving the underlying behavior. A desirable reward should respond to the former while remaining stable to the latter. We further examine how the DiTaR reward evolves during policy learning.

\subsection{Temporal Sensitivity on Bridge and Real-World Videos}
\label{sec:suppl_reward_temporal}

We apply backward drift, looped segments, shuffled chunks, and duplicate frames to expert videos. These perturbations alter the sequence without changing the appearance of the individual frames. Shuffled chunks disrupt the order of execution, while duplicate frames introduce repetition without new state changes. The comparison tests whether a reward responds to these temporal changes even though the observations still come from expert videos.

\begin{table}[!htb]
    \centering
    \footnotesize
    \setlength{\tabcolsep}{5pt}
    \renewcommand{\arraystretch}{1.05}
    \caption{\textbf{Temporal sensitivity.} Signed reward changes $\Delta$ (\%) over 16 expert episodes per dataset. More negative values indicate stronger temporal sensitivity. The largest decrease per row and dataset is bold. Diff. R.: Diffusion Reward.}
    \label{tab:reward_corruption}
    \vspace{2mm}
    \begin{tabular}{lccc@{\hspace{14pt}}ccc}
        \toprule
        & \multicolumn{3}{c}{\textbf{Bridge}} & \multicolumn{3}{c}{\textbf{Real-world}} \\
        \cmidrule(lr){2-4}\cmidrule(l){5-7}
        Corruption & DiTaR & Diff. R. & VIPER & DiTaR & Diff. R. & VIPER \\
        \midrule
        \multicolumn{7}{l}{\textit{Signed reward change: $\Delta$ (\%) $\downarrow$}} \\
        Backward drift & $\mathbf{-2.4}$ & $-1.1$ & $-0.5$ & $\mathbf{-1.6}$ & $+7.0$ & $+0.6$ \\
        Looped segment & $\mathbf{-3.5}$ & $-0.6$ & $-1.7$ & $\mathbf{-1.0}$ & $+7.4$ & $+0.7$ \\
        Shuffled chunks & $\mathbf{-16.8}$ & $-0.7$ & $-6.0$ & $-9.2$ & $-2.5$ & $\mathbf{-11.8}$ \\
        Duplicate frames & $\mathbf{-21.1}$ & $+0.9$ & $+7.0$ & $\mathbf{-26.4}$ & $+6.8$ & $+5.3$ \\
        \bottomrule
    \end{tabular}
\end{table}

Table~\ref{tab:reward_corruption} shows that DiTaR's reward decreases under every corruption in both datasets. On Bridge, the decreases are largest for shuffled chunks and duplicate frames, at $-16.8\%$ and $-21.1\%$, respectively. Backward drift and looped segments produce smaller decreases. On real-world videos, DiTaR also gives a lower reward in all four cases, although VIPER shows a larger decrease for shuffled chunks.

\begin{figure}[!htb]
    \centering
    \includegraphics[width=0.98\linewidth]{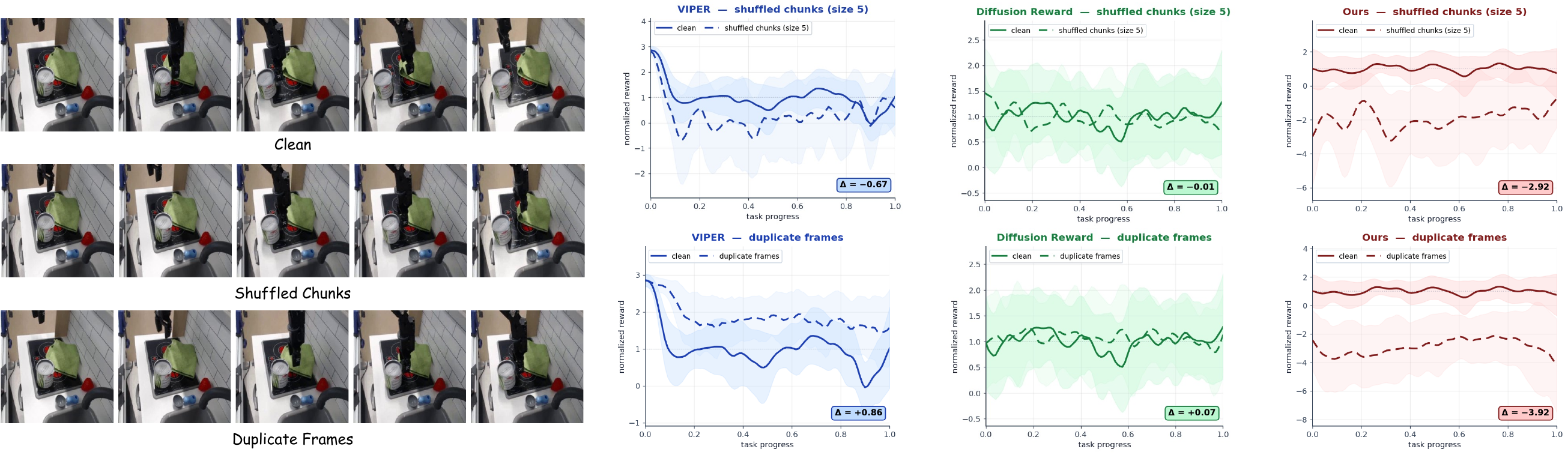}
    \caption{\textbf{Temporal corruption on Bridge.} Clean and corrupted rewards over 16 episodes. Bands indicate one standard deviation; $\Delta$ is the mean signed reward shift.}
    \label{fig:suppl_reward_corruption}
\end{figure}

Fig.~\ref{fig:suppl_reward_corruption} shows how these differences appear over time. DiTaR assigns lower rewards to both corrupted sequences across much of the episode. VIPER also responds to shuffled chunks, but its reward increases under frame duplication; Diffusion Reward changes little in either case. Frame duplication is informative because it preserves the appearance of expert observations but removes continued state changes. DiTaR's lower reward reflects its sensitivity to these missing changes.

\subsection{Temporal Corruption Localization}
\label{sec:suppl_reward_localization}

A decrease in the mean reward does not indicate when the model responds to a temporal perturbation. We therefore apply temporal jitter or local shuffling only to the middle third of each video, leaving the remaining frames unchanged. Localization AUC measures whether the clean-to-corrupted reward decrease distinguishes affected from unaffected output positions. For DiTaR, affected positions include clips whose temporal support overlaps the corrupted interval.

\begin{figure}[!htb]
    \centering
    \includegraphics[width=\linewidth]{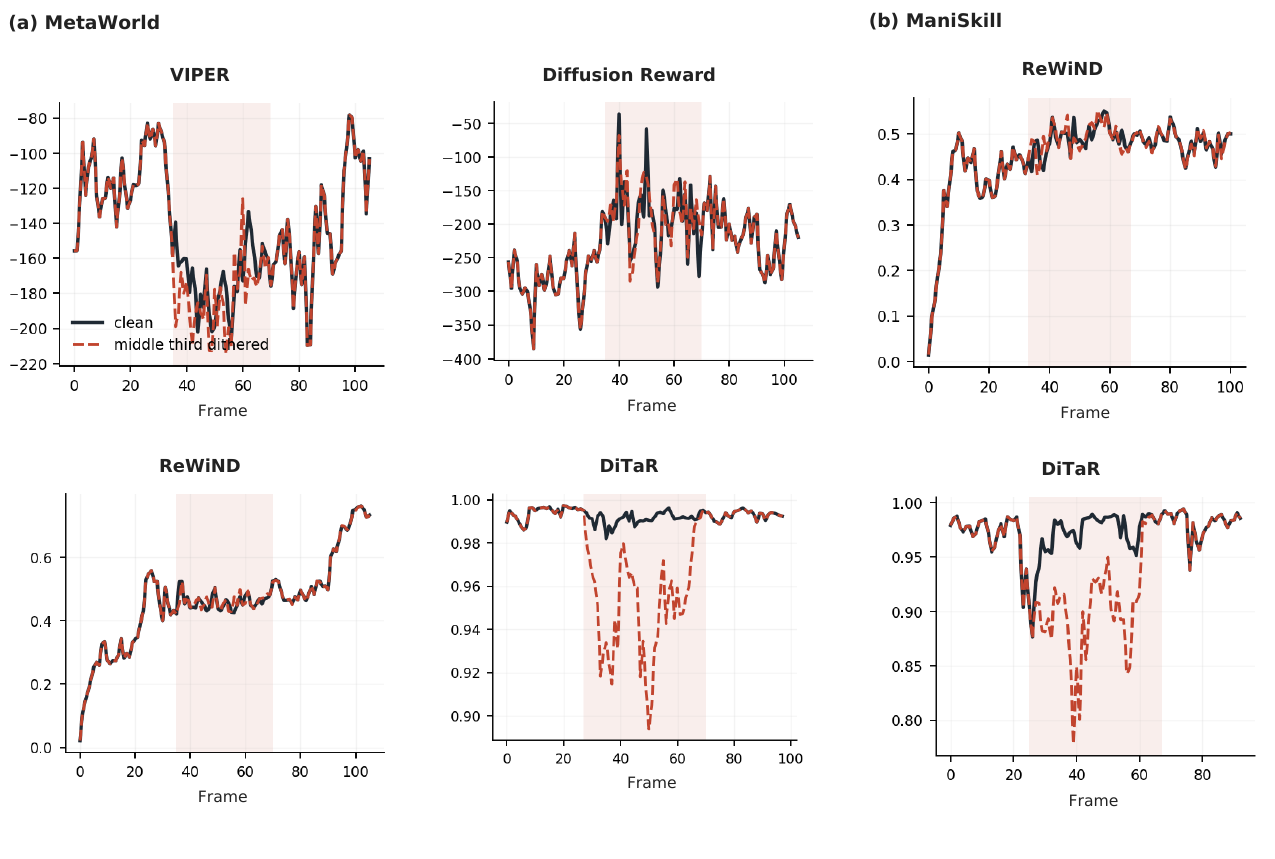}
    \caption{\textbf{Temporal corruption localization.} Temporal jitter of $\pm2$ frames is applied to the shaded middle third. Left two columns: MetaWorld, ``take the object out of the hole.'' Right: ManiSkill, ``push the red T-shaped block into place.'' Solid and dashed curves show clean and corrupted rewards.}
    \label{fig:suppl_reward_localization}
\end{figure}

Table~\ref{tab:suppl_localization_auc} reports mean localization AUCs. DiTaR localizes both perturbations on MetaWorld and ManiSkill, with higher AUCs for local shuffling than for temporal jitter in both datasets. VIPER also localizes the MetaWorld corruptions. These results measure whether reward changes identify the temporally affected interval. Fig.~\ref{fig:suppl_reward_localization} shows representative examples from each dataset.

\begin{table}[!htb]
    \centering
    \small
    \setlength{\tabcolsep}{5pt}
    \renewcommand{\arraystretch}{1.05}
    \caption{\textbf{Temporal corruption localization.} Mean localization AUCs on MetaWorld and ManiSkill; higher values indicate better localization.}
    \label{tab:suppl_localization_auc}
    \vspace{2mm}
    \begin{tabular}{llcc}
        \toprule
        Dataset & Method & Temporal jitter $\uparrow$ & Local shuffling $\uparrow$ \\
        \midrule
        MetaWorld & VIPER & 0.91 & 0.96 \\
        MetaWorld & DiTaR & 0.95 & 1.00 \\
        ManiSkill & DiTaR & 0.83 & 0.94 \\
        \bottomrule
    \end{tabular}
\end{table}

In the MetaWorld example, DiTaR's clean reward remains high through the middle third, while temporal jitter introduces several distinct decreases within the perturbed interval. The clean and corrupted curves largely coincide before the perturbation and approach one another again afterward. VIPER also responds near the altered segment, while Diffusion Reward and ReWiND show smaller differences. The localization AUCs for this video are 0.76, 0.47, 0.39, and 0.98 for VIPER, Diffusion Reward, ReWiND, and DiTaR, respectively.

The ManiSkill example shows the same behavior. DiTaR's reward decreases near the perturbed interval and recovers afterward, whereas the two ReWiND curves remain similar throughout. Their localization AUCs are 0.94 and 0.46, respectively.

\subsection{Visual Robustness}
\label{sec:suppl_reward_viewpoint}

\paragraph{Viewpoint changes.}
We test whether viewpoint changes alter the DiTaR reward when the underlying task execution is unchanged. On ManiSkill, we apply synthetic camera yaw angles of $4^\circ$, $8^\circ$, $16^\circ$, $24^\circ$, and $32^\circ$, extending the fixed-angle comparison in the main paper. We also compare two rendered camera views of the same rollout.

\begin{figure}[!htb]
    \centering
    \includegraphics[width=0.76\linewidth]{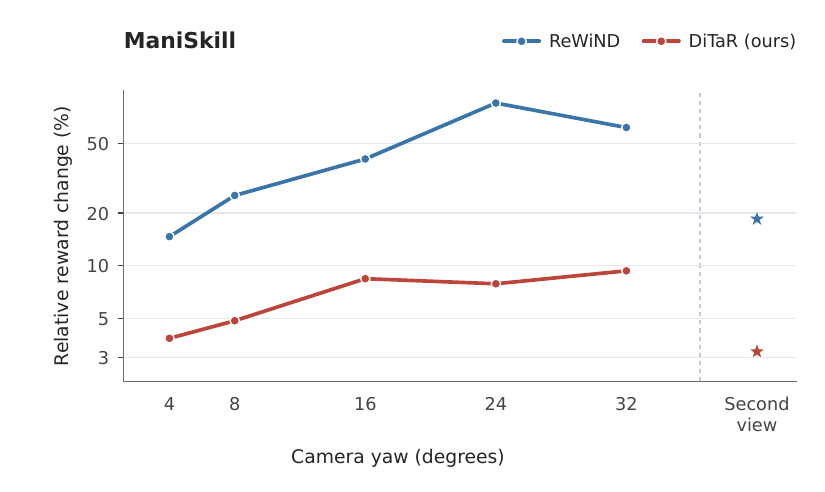}
    \caption{\textbf{Viewpoint robustness on ManiSkill.} Relative reward changes under synthetic camera yaw (circles) and a second rendered view of the same rollout (stars). The vertical axis is logarithmic; smaller changes indicate greater robustness.}
    \label{fig:suppl_reward_viewpoint}
\end{figure}

Fig.~\ref{fig:suppl_reward_viewpoint} shows smaller relative reward changes for DiTaR than ReWiND throughout the tested yaw range. ReWiND's change is $10.7\times$ that of DiTaR at $24^\circ$ and $6.6\times$ at $32^\circ$. Table~\ref{tab:suppl_viewpoint_summary} further shows that DiTaR has the smaller change in 82--86\% of videos at the three smaller angles and in all evaluated videos at the two larger angles.

\begin{table}[!htb]
    \centering
    \footnotesize
    \setlength{\tabcolsep}{5pt}
    \renewcommand{\arraystretch}{1.05}
    \caption{\textbf{Viewpoint robustness on ManiSkill.} Relative reward-change ratio (ReWiND/DiTaR) and percentage of videos with a smaller reward change for DiTaR.}
    \label{tab:suppl_viewpoint_summary}
    \begin{tabular}{lccccc}
        \toprule
        Camera yaw & $4^\circ$ & $8^\circ$ & $16^\circ$ & $24^\circ$ & $32^\circ$ \\
        \midrule
        Relative change ratio & $3.8\times$ & $5.2\times$ & $4.8\times$ & $10.7\times$ & $6.6\times$ \\
        Smaller DiTaR change (\% of videos) & 82 & 86 & 86 & 100 & 100 \\
        \bottomrule
    \end{tabular}
\end{table}

For the second-camera comparison, the aggregate reward changes are 3.2\% for DiTaR and 18.5\% for ReWiND, further showing that DiTaR is less affected when the same execution is observed from a different viewpoint.

\begin{figure}[!htb]
    \centering
    \includegraphics[width=0.90\linewidth]{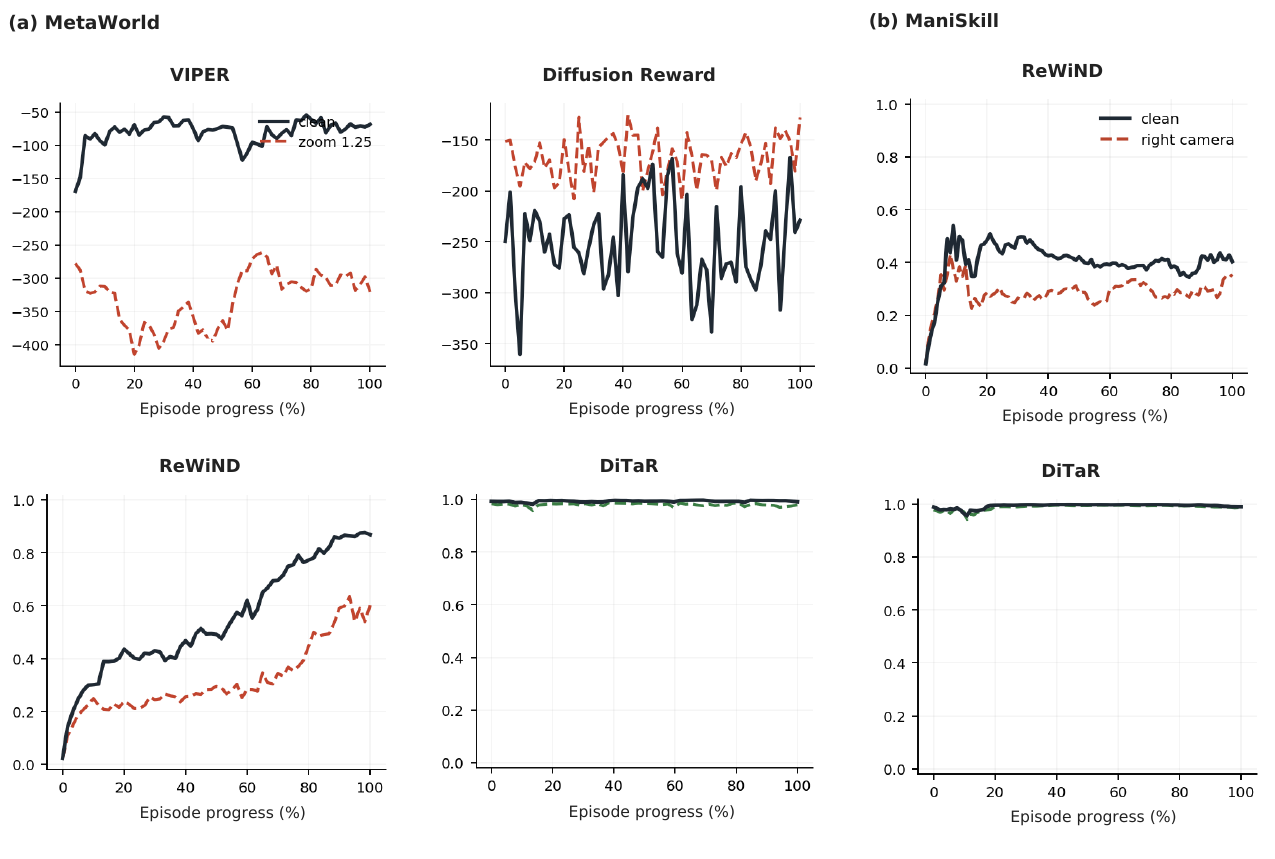}
    \caption{\textbf{Rewards under visual perturbations.} Left two columns: MetaWorld, ``press the button,'' under $1.25\times$ camera zoom. Right column: ManiSkill, ``push the red T-shaped block into place,'' under a second camera view. Solid and dashed curves show clean and perturbed rewards.}
    \label{fig:suppl_reward_visual_trajectories}
\end{figure}

\paragraph{Reward trajectories.}
\label{sec:suppl_reward_visual_trajectories}
Fig.~\ref{fig:suppl_reward_visual_trajectories} compares complete reward trajectories for a MetaWorld camera-zoom example and a ManiSkill second-view example.

Full reward trajectories also show how visual perturbations affect the reward over time. In the MetaWorld example, camera zoom shifts VIPER's reward downward and Diffusion Reward's reward upward across much of the episode. ReWiND retains an increasing trend, but its perturbed reward remains below the clean reward. DiTaR's two trajectories largely overlap. The relative mean reward changes are 314.1\%, 33.5\%, 42.2\%, and 1.4\% for VIPER, Diffusion Reward, ReWiND, and DiTaR.

In the ManiSkill example, switching cameras lowers ReWiND's reward over most of the rollout, while DiTaR preserves a similar trajectory. Their relative mean reward changes are 29.0\% and 0.4\%, respectively. Together with the localization results, these examples show that DiTaR responds to temporal disruptions while remaining comparatively stable when the same execution is viewed from another camera.

\subsection{Reward Evolution during Policy Learning}
\label{sec:suppl_reward_training}

We next examine how the DiTaR reward evolves as policy learning progresses. For each task, we compare rewards at the same task-specific timestep across different stages of training.

\begin{figure}[!htb]
    \centering
    \includegraphics[width=0.95\linewidth]{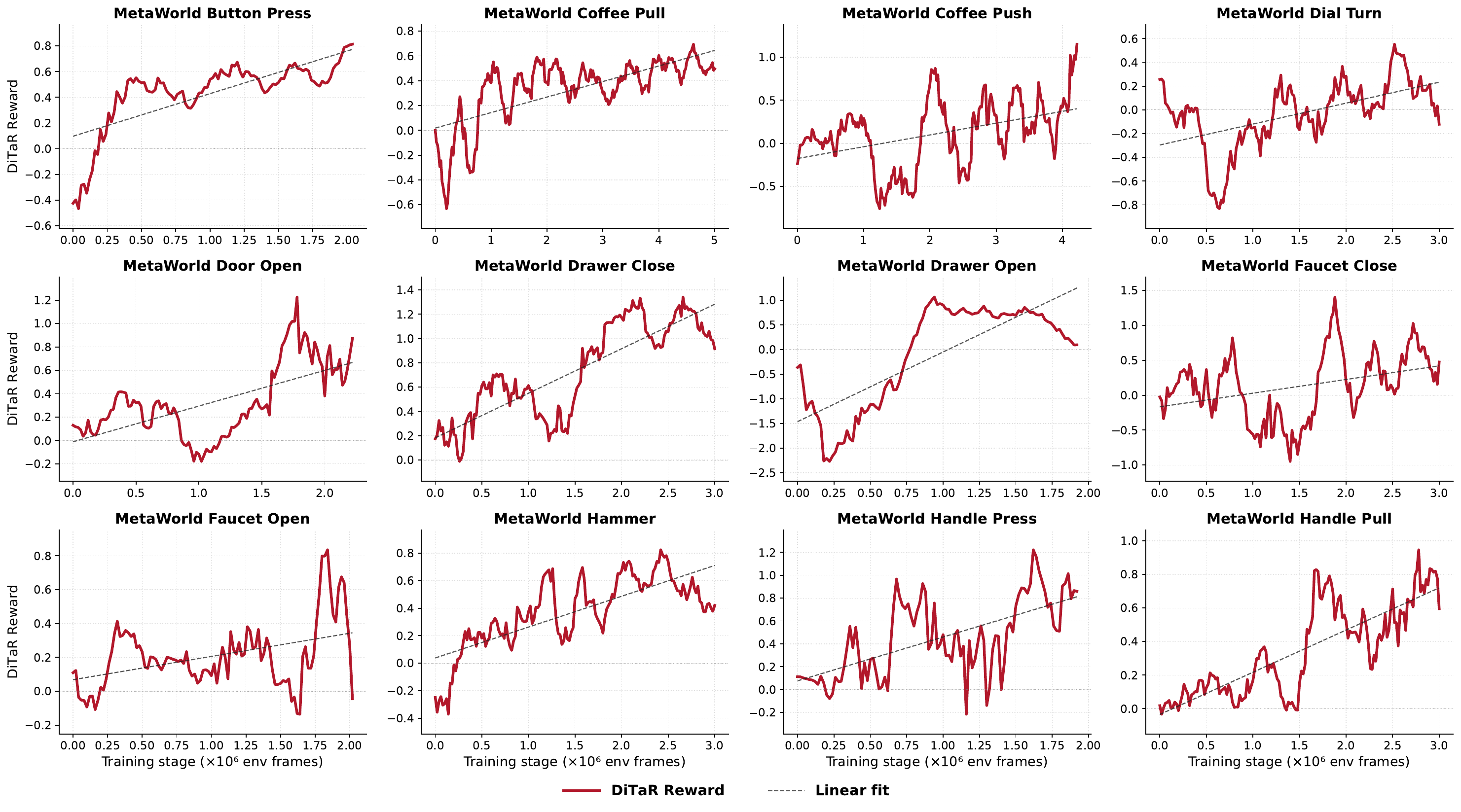}
    \caption{\textbf{Reward evolution during policy learning.} DiTaR rewards at a fixed task-specific timestep across training stages on 12 MetaWorld-v3 tasks. Solid curves show smoothed traces and dashed lines show linear fits.}
    \label{fig:suppl_reward_progression}
\end{figure}

Fig.~\ref{fig:suppl_reward_progression} shows an overall positive trend on all 12 MetaWorld-v3 tasks. Although the reward is not monotonic on every task, later stages generally receive higher rewards. These results complement the perturbation analyses by showing that the DiTaR reward also tracks improvements in policy behavior during training.

\section{Ablation and Latent Analysis}
\label{sec:suppl_ablation_latent}

\subsection{Task-Relevant Latent Ablation}
\label{sec:supp_tr_latent_ablation}

We compare DiTaR's task-relevant latent encoding with a single-stage reconstruction-based encoder trained without language conditioning. The ablated encoder compresses frozen Wan VAE latents, and its output replaces $z_t^{\mathrm{rel}}$ as the target for dynamics modeling. We retrain the history- and language-conditioned rectified-flow model on these latents using the settings in Table~\ref{tab:suppl_hparam_diffusion}. The TD reward formulation, reward normalization, and downstream RL settings remain unchanged.

\begin{figure}[!htb]
    \centering
    \includegraphics[width=0.89\textwidth]{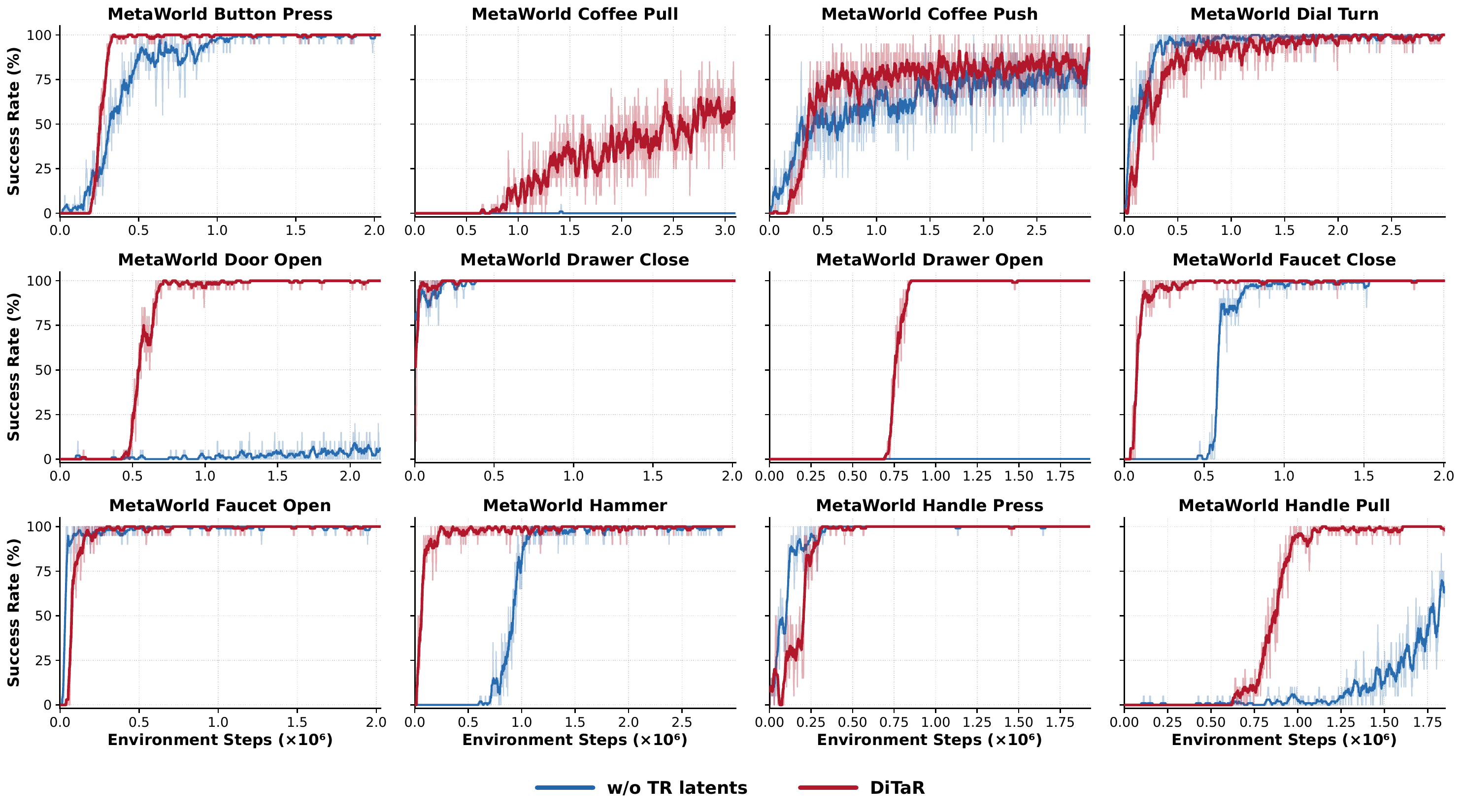}
    \caption{\textbf{Task-relevant latent ablation.} Learning curves on 12 MetaWorld-v3 tasks comparing DiTaR's task-relevant latents with latents from a single-stage reconstruction-based encoder trained without language conditioning. Both variants use the same conditional rectified-flow architecture and TD reward formulation, with the dynamics model trained separately for each latent representation. DiTaR achieves higher overall success and learns more efficiently.}
    \label{fig:tr_latent_ablation}
\end{figure}

Table~\ref{tab:wan_ablation_aggregate} shows that DiTaR improves final success from 67.2\% to 94.3\% and reduces NormT$@$0.8 from 0.458 to 0.244. SR$\geq$80 increases from 58.3\% to 91.7\%. The per-task curves in Fig.~\ref{fig:tr_latent_ablation} show substantial differences on Coffee Pull, Drawer Open, and Handle Pull, where the reconstruction-based variant learns more slowly or does not reach high success within the plotted interval. These results support using the task-relevant representation for reward modeling.

\begin{table}[!htb]
\centering
\small
\caption{Task-relevant latent ablation on MetaWorld-v3, with baseline results for comparison. Metrics are defined in Sec.~\ref{sec:suppl_evaluation_metrics}.}
\label{tab:wan_ablation_aggregate}
\begin{tabular}{lcccc}
\toprule
Method & Final SR (\%) $\uparrow$ & SR$\ge$80 (\%) $\uparrow$ & Reach 0.8 (\%) $\uparrow$ & NormT@0.8 $\downarrow$ \\
\midrule
Env-only & 56.8 & 50.0 & 50.0 & 0.678 \\
VIPER & 39.9 & 33.3 & 41.7 & 0.734 \\
Diffusion Reward & 13.1 & 8.3 & 8.3 & 0.922 \\
ReWiND & 65.2 & 33.3 & 41.7 & 0.609 \\
\midrule
w/o TR latents & 67.2 & 58.3 & 66.7 & 0.458 \\
\textbf{DiTaR} & \textbf{94.3} & \textbf{91.7} & \textbf{91.7} & \textbf{0.244} \\
\bottomrule
\end{tabular}
\end{table}

\subsection{History and Temporal-Difference Ablations}
\label{sec:suppl_history_td_ablation}

\paragraph{Without history conditioning.}
We remove the encoded five-frame history from the dynamics model while retaining the task-relevant target, language conditioning, architecture size, training objective, and optimization schedule. Reward inference, normalization, and policy training otherwise follow the full model. The learning curves are shown in Fig.~\ref{fig:ablation_history} of the main paper.

\paragraph{Without temporal differencing.}
We keep the latent encoders and conditional dynamics model fixed and replace $r_t^{\mathrm{TD}}=\phi_t-\phi_{t-\Delta t}$ with $r_t^{\mathrm{nonTD}}=\phi_t=\exp(-e_t)$. All other settings are unchanged. This comparison tests the use of temporal changes in compatibility against its absolute value; the results are shown in Fig.~\ref{fig:ablation_td} of the main paper.

\subsection{Cross-Dataset Latent Grouping}
\label{sec:latent_analysis}

\begin{figure}[!htb]
    \centering
    \includegraphics[width=0.89\textwidth]{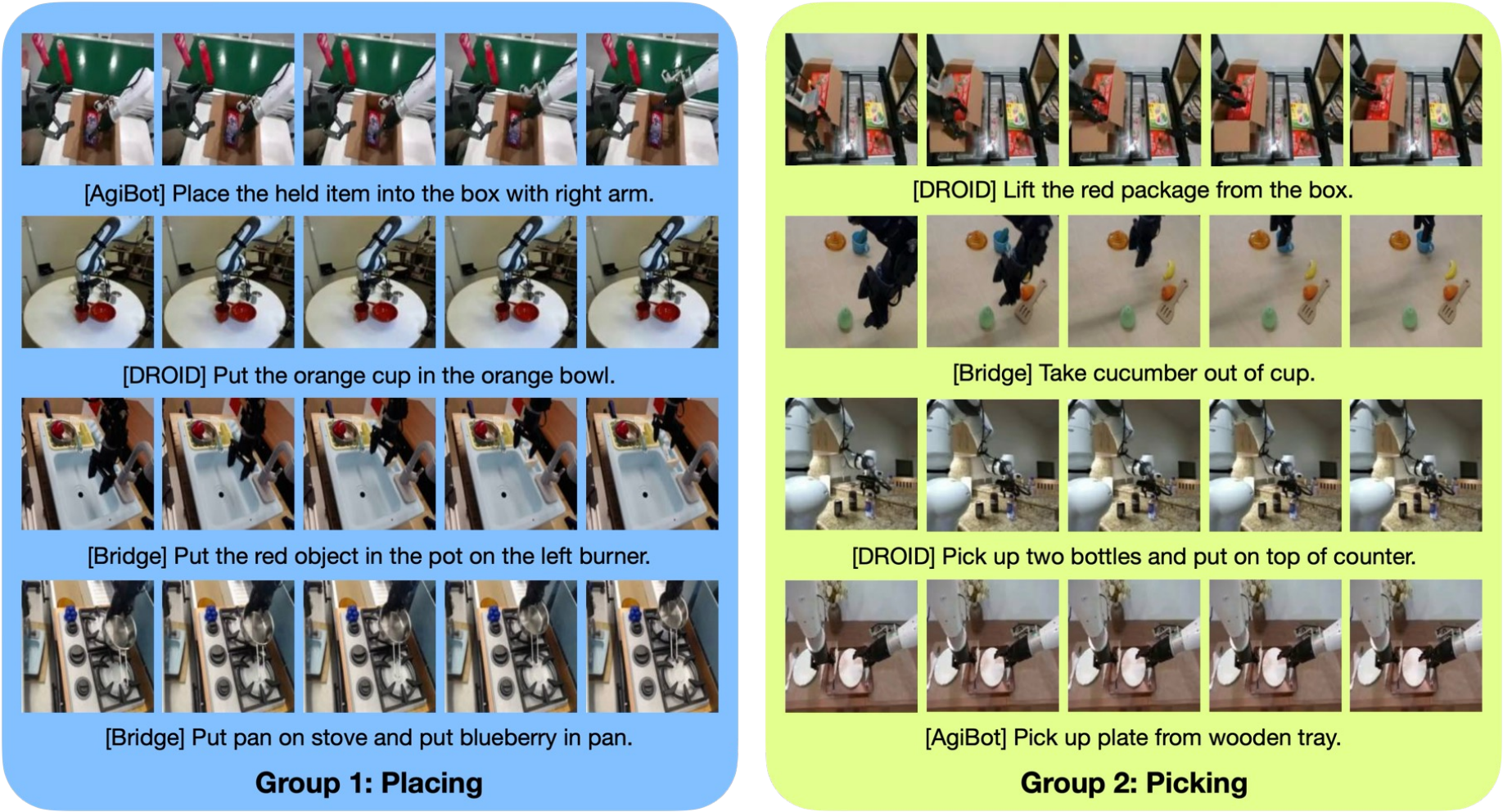}
    \caption{\textbf{Cross-dataset latent grouping.} Groups of five-frame clips with pairwise cosine similarity of at least 0.90 in the task-relevant latent space. The left and right groups show placing and picking behaviors across different robots and visual settings.}
    \label{fig:suppl_latent_grouping}
\end{figure}

For each five-frame clip, we mean-pool its task-relevant latent over the temporal and query axes, $\ell_2$-normalize the resulting descriptor, and compute cosine similarity between clips. Each group in Fig.~\ref{fig:suppl_latent_grouping} contains only clips with pairwise similarity of at least 0.90. The groups show placing and picking behaviors across datasets, with different robots, viewpoints, backgrounds, and objects. These examples illustrate shared manipulation behaviors in the learned latent space.

\section{Model and Reward Implementation}
\label{sec:suppl_impl}

\subsection{Task-Relevant Latent Encoder}
\label{sec:suppl_latent_encoder}
We use the frozen Wan-2.2 TI2V-5B VAE~\citep{wan2025wan} to encode each strided five-frame clip into video latents $V=(V_0,V_1)$. The encoders $E_{\mathrm{irr}}$ and $E_{\mathrm{rel}}$ have separate learnable queries and KL-regularized variational bottlenecks. Task instructions are embedded by a frozen T5-base encoder~\citep{raffel2020exploring}.

In Stage 1, we train $E_{\mathrm{irr}}$ and decoder $D_\psi$ to reconstruct $V_1$ from $V_0$, $z^{\mathrm{irr}}$, and the language embedding $e_\ell$. In Stage 2, both $E_{\mathrm{irr}}$ and $D_\psi$ remain frozen. We train only $E_{\mathrm{rel}}$, which takes $V$ and $z^{\mathrm{irr}}$ as input and produces $z^{\mathrm{rel}}$. The fixed decoder receives $z^{\mathrm{rel}}$ in place of its direct language input. With the Stage 1 pathway fixed, $E_{\mathrm{rel}}$ learns to recover from video the task-relevant information previously supplied to the decoder through language. The bottleneck tokens are projected to the decoder conditioning dimension before reconstruction. Table~\ref{tab:suppl_hparam_stage2} lists the architecture and training settings.

\begin{table}[!htb]
    \centering
    \footnotesize
    \setlength{\tabcolsep}{5pt}
    \renewcommand{\arraystretch}{1.0}
    \caption{Architecture and training settings for the latent encoders. The pretrained Wan VAE and T5-base text encoder are frozen throughout representation learning.}
    \label{tab:suppl_hparam_stage2}
    \begin{tabular}{@{}p{0.32\linewidth}p{0.55\linewidth}@{}}
        \toprule
        Hyperparameter & Value \\
        \midrule
        Input latent dimension & 48 \\
        Frames per training window & 5 \\
        Model dimension & 768 \\
        Attention heads & 12 \\
        Encoder blocks & 4 \\
        Decoder blocks & 12 \\
        Query blocks & 8 \\
        Query tokens & 4 \\
        Bottleneck dimension & 32 \\
        Bottleneck regularizer & KL \\
        Separate task-relevant encoder & Yes \\
        Text encoder & Frozen T5-base; dimension 768; maximum length 64 \\
        Optimizer & AdamW \\
        Batch size & 128 per GPU \\
        Training steps & $\sim\!1.0\times10^{4}$ \\
        Training cost & $8$ GPUs $\times\sim\!6$ hours $=\sim\!48$ GPU-hours \\
        \bottomrule
    \end{tabular}
\end{table}

\subsection{Conditional Rectified-Flow Model}
\label{sec:suppl_dynamics_model}

The dynamics model $f_\theta$ predicts the clean task-relevant latent from its noisy input, the visual history, and the task instruction. A five-frame history is encoded by the frozen Wan VAE into two latent frames and flattened into $2\times256=512$ history tokens. History and target tokens are separately projected to the common model dimension ($d=512$) before being concatenated for self-attention in each DiT block. Language conditioning is applied to target tokens through gated cross-attention, with the scalar gate initialized to zero. The Wan VAE and both latent encoders remain frozen during dynamics training. Table~\ref{tab:suppl_hparam_diffusion} reports the model settings.

\begin{table}[!htb]
    \centering
    \footnotesize
    \setlength{\tabcolsep}{5pt}
    \renewcommand{\arraystretch}{1.0}
    \caption{Architecture and training settings for the conditional rectified-flow model.}
    \label{tab:suppl_hparam_diffusion}
    \begin{tabular}{@{}p{0.32\linewidth}p{0.55\linewidth}@{}}
        \toprule
        Hyperparameter & Value \\
        \midrule
        History input & Five-frame history encoded by the Wan VAE \\
        History tokens / token dimension & 512 / 48 \\
        Target input & Task-relevant latent of the target clip \\
        Target tokens / token dimension & 4 / 32 \\
        Text conditioning & Target-only cross-attention; scalar gate initialized to 0 \\
        Model dimension & 512 \\
        DiT blocks & 12 \\
        Attention heads & 8 \\
        Timestep conditioning & Sinusoidal embedding and AdaLN modulation \\
        Training objective & Endpoint regression \\
        History dropout / text dropout & 0.0 / 0.15 \\
        Optimizer & AdamW \\
        Batch size & 128 per GPU \\
        Training steps & $\sim\!6\times10^{4}$ \\
        Training cost & $8$ GPUs $\times\sim\!24$ hours $=\sim\!192$ GPU-hours \\
        \bottomrule
    \end{tabular}
\end{table}

\subsection{Reward Computation}
\label{sec:suppl_reward_computation}

For each observed target clip, we extract $z_t^{\mathrm{rel}}$ and encode its preceding history as $H_t=E_{\mathrm{Wan}}(h_t)$. We evaluate the endpoint residual using the frozen dynamics model,
\begin{equation}
    e_t=\mathbb{E}_{\tau,\epsilon}\!\left[\frac{1}{d_z}\left\|f_\theta(z_{\tau,t},H_t,e_\ell,\tau)-z_t^{\mathrm{rel}}\right\|_2^2\right],
    \qquad z_{\tau,t}=(1-\tau)z_t^{\mathrm{rel}}+\tau\epsilon,
    \label{eq:suppl_endpoint_residual}
\end{equation}
where $\epsilon\sim\mathcal{N}(0,I)$ and $\tau\sim\mathcal{U}(\tau_{\min},\tau_{\max})$. The compatibility score and TD reward are
\begin{equation}
    \phi_t=\exp(-e_t),\qquad r_t^{\mathrm{TD}}=\phi_t-\phi_{t-\Delta t}.
    \label{eq:suppl_td_reward}
\end{equation}
We compute reward normalization statistics before policy training and use $\widetilde r_t^{\mathrm{TD}}=(r_t^{\mathrm{TD}}-\mu_D)/\sigma_D$ in the RL objective. The temporal gap is $\Delta t=4$. We use two Monte Carlo samples to estimate the expectation in Eq.~\eqref{eq:suppl_endpoint_residual}, with the same sampled $(\tau,\epsilon)$ pairs used for $\phi_t$ and $\phi_{t-\Delta t}$.

\section{Policy Learning and Evaluation}
\label{sec:suppl_policy_evaluation}

\subsection{Online Visual RL}
\label{sec:suppl_impl_rl}

Our online RL experiments use the DrQ-v2~\citep{yaratsmastering} pipeline from Diffusion Reward~\citep{huang2024diffusionreward}. The policy and critic architectures, observation size, image augmentation, replay buffer, exploration schedule, RND module, and environment interface are shared across all methods. In particular, Env-only and all learned-reward baselines use the same RND exploration setup; during policy learning, the learned-reward methods differ only in the learned reward signal. Learned rewards are computed before transitions enter the replay buffer.

The policy is trained with
\begin{equation}
    R_t=\lambda_D\widetilde r_t^{\,\mathrm{learn}}+r_t^{\mathrm{env}}+\lambda_{\mathrm{expl}}r_t^{\mathrm{RND}},
    \label{eq:suppl_rl_reward}
\end{equation}
where $\widetilde r_t^{\,\mathrm{learn}}$ is the normalized learned reward, $r_t^{\mathrm{env}}$ is the sparse benchmark reward, and $r_t^{\mathrm{RND}}$ is the RND exploration bonus. VIPER, Diffusion Reward, ReWiND, and DiTaR use the same reward-normalization procedure: normalization statistics are computed before policy learning and kept fixed throughout training. DiTaR uses $\widetilde r_t^{\mathrm{TD}}$ for the learned term. For learned-reward methods, the task-specific $(\lambda_D,\lambda_{\mathrm{expl}})$ pairs are shared across methods; Env-only sets $\lambda_D=0$ while retaining the same RND exploration setting. Tables~\ref{tab:suppl_hparam_rl} and~\ref{tab:suppl_reward_mix_scales} list the RL settings and reward coefficients, respectively. Algorithm~\ref{alg:suppl_online_rl} summarizes their use during training.

\begin{table}[!htb]
    \centering
    \footnotesize
    \setlength{\tabcolsep}{5pt}
    \renewcommand{\arraystretch}{1.0}
    \caption{Online visual RL settings for MetaWorld-v3 and ManiSkill.}
    \label{tab:suppl_hparam_rl}
    \begin{tabular}{@{}p{0.41\linewidth}p{0.47\linewidth}@{}}
        \toprule
        Hyperparameter & Value \\
        \midrule
        \multicolumn{2}{@{}l}{\textbf{Environment}} \\
        Action repeat & 1 \\
        Frame stack & 1 \\
        Observation size & $64\times64\times3$ \\
        Episode length & 250 environment steps \\
        Per-task training budget & $3\times10^6$ environment frames \\
        \midrule
        \multicolumn{2}{@{}l}{\textbf{DrQ-v2}} \\
        Data augmentation & Random shift, $\pm4$ pixels \\
        Replay buffer capacity & $10^6$ \\
        Discount $\gamma$ & 0.99 \\
        $n$-step returns & 3 \\
        Initial exploration frames & 4{,}000 \\
        Feature dimension & 50 \\
        Hidden dimension & 1024 \\
        Exploration standard-deviation clip & 0.3 \\
        Exploration standard-deviation schedule & Linear$(1.0,0.1,3\times10^6)$ \\
        Soft update rate & 0.01 \\
        Optimizer & Adam, learning rate $10^{-4}$ \\
        Batch size & 256 \\
        Update frequency & 2 \\
        \midrule
        \multicolumn{2}{@{}l}{\textbf{RND exploration}} \\
        CNN feature dimension & $7\times7\times64$ \\
        MLP size & $512\times512$ \\
        Learning rate & $10^{-4}$ \\
        \midrule
        \multicolumn{2}{@{}l}{\textbf{DiTaR reward}} \\
        Reward formulation & TD flow reward \\
        Temporal gap $\Delta t$ & 4 \\
        Paired Monte Carlo samples & 2 \\
        Sampling steps & 16 \\
        Reward and exploration coefficients & Table~\ref{tab:suppl_reward_mix_scales} \\
        \bottomrule
    \end{tabular}
\end{table}

\begin{table}[!htb]
    \centering
    \scriptsize
    \setlength{\tabcolsep}{3pt}
    \renewcommand{\arraystretch}{1.0}
    \setlength{\tabcolsep}{4pt}
    \caption{Task-specific reward and exploration coefficients $(\lambda_D,\lambda_{\mathrm{expl}})$ in Eq.~\eqref{eq:suppl_rl_reward}, shared across learned-reward methods. Following the reward-mixing setup of Diffusion Reward~\citep{huang2024diffusionreward}, we use task-specific reward and exploration coefficients. Tasks sharing the same coefficient pair are grouped in a single row.}
    \label{tab:suppl_reward_mix_scales}
    \begin{tabular}{llp{0.60\linewidth}}
        \toprule
        Benchmark & $(\lambda_D,\lambda_{\mathrm{expl}})$ & Tasks \\
        \midrule
        MetaWorld-v3 & $(0.05,0.95)$ & \texttt{button-press}, \texttt{coffee-pull}, \texttt{coffee-push}, \texttt{door-open}, \texttt{drawer-open}, \texttt{faucet-close}, \texttt{faucet-open}, \texttt{hammer}, \texttt{handle-press}, \texttt{handle-pull} \\
        & $(0.95,0.05)$ & \texttt{drawer-close} \\
        & $(0.50,0.50)$ & \texttt{dial-turn} \\
        \midrule
        ManiSkill & $(0.05,0.95)$ & \texttt{PullCube}, \texttt{PushCube} \\
        & $(0.95,0.05)$ & \texttt{PokeCube} \\
        & $(0.50,0.05)$ & \texttt{PullCubeTool} \\
        \bottomrule
    \end{tabular}
\end{table}

\subsection{Reward Baselines}
\label{sec:suppl_impl_baselines}

VIPER, Diffusion Reward, and ReWiND are trained on expert videos from the corresponding MetaWorld or ManiSkill tasks. We reproduce the reward models for all learned-reward baselines following their official released implementations. All methods use the online RL pipeline described above.

\paragraph{Env-only.}
We set $\lambda_D=0$, retaining the sparse benchmark reward and RND exploration bonus without a learned reward model.

\paragraph{VIPER.}
We retrain VideoGPT on the task-specific expert videos and use conditional video likelihood as reward~\citep{escontrela2023viper}. Reward inference follows the original teacher-forcing implementation and uses the original model hyperparameters.

\paragraph{Diffusion Reward.}
We retrain VQ-Diffusion on the same expert videos used for VIPER and use its history-conditioned uncertainty reward~\citep{huang2024diffusionreward}. We retain the original reward formulation and evaluate it in the shared RL pipeline.

\paragraph{ReWiND.}
We train a language-conditioned progress estimator on the corresponding expert videos, using temporal transformations to construct progress supervision~\citep{zhang2025rewind}. The resulting learned reward is evaluated in the same DrQ-v2 pipeline.

\subsection{Evaluation Metrics}
\label{sec:suppl_evaluation_metrics}

We use the aggregate metrics defined in the main paper. Final SR is the mean final success rate. SR$\geq$80 is the percentage of tasks with final success rate at least 80\%. Reach 0.8 is the percentage of tasks whose rolling-window smoothed success rate reaches 0.8 at least once during training. NormT@0.8 is the mean first hitting time divided by the corresponding training budget, assigning a value of 1 when the threshold is not reached.

\subsection{Real-World Offline RL}
\label{sec:suppl_impl_offline}

The real-world evaluation uses 150 demonstrations collected with a Franka Research 3 robot across the three tasks shown in Fig.~\ref{fig:real_world_demo}. The recordings contain synchronized third-person and wrist RGB observations. For each reward model, we relabel the same offline transitions and train a policy with Implicit Q-Learning (IQL)~\citep{kostrikov2022offline}. The demonstrations, policy architecture, and IQL settings are shared across reward models. We conduct 10 evaluation trials per task; results are reported in Table~\ref{tab:real_offline_rl} of the main paper.

\section{Algorithms}
\label{sec:suppl_algorithms}

Algorithm~\ref{alg:ditar} summarizes the two-stage representation learning and conditional dynamics training. Algorithm~\ref{alg:suppl_online_rl} shows how the frozen reward model is used during online policy learning.

\begin{algorithm}[!htbp]
    \caption{DiTaR Representation and Dynamics Training}
    \label{alg:ditar}
    \begin{algorithmic}[1]
    \footnotesize
    \Require Video--instruction dataset $\mathcal{D}$; frozen $E_{\mathrm{Wan}}$ and $f_{\mathrm{text}}$
    \Ensure Trained $E_{\mathrm{irr}}$, $E_{\mathrm{rel}}$, $f_\theta$; reward statistics $(\mu_D,\sigma_D)$
    \Statex \textbf{Stage 1: task-irrelevant latent encoding}
    \State Initialize $E_{\mathrm{irr}}$ and reconstruction decoder $D_\psi$
    \While{not converged}
        \State Sample clip and instruction $(x,\ell)$ from $\mathcal{D}$
        \State $V=(V_0,V_1)\gets E_{\mathrm{Wan}}(x)$; $e_\ell\gets f_{\mathrm{text}}(\ell)$
        \State Sample $z^{\mathrm{irr}}\sim q_{\eta_{\mathrm{irr}}}(\cdot\mid V,e_\ell)$
        \State $\widehat V_1^{(1)}\gets V_0+D_\psi(V_0,z^{\mathrm{irr}},e_\ell)$
        \State Update $E_{\mathrm{irr}}$ and $D_\psi$ using $\mathcal{L}_{\mathrm{irr}}$
    \EndWhile
    \Statex \textbf{Stage 2: task-relevant latent encoding}
    \State Freeze $E_{\mathrm{irr}}$ and $D_\psi$; initialize $E_{\mathrm{rel}}$
    \While{not converged}
        \State Sample $(x,\ell)$; compute $V=(V_0,V_1)$ and $e_\ell$
        \State Extract $z^{\mathrm{irr}}$ with the frozen $E_{\mathrm{irr}}$
        \State Sample $z^{\mathrm{rel}}\sim q_{\eta_{\mathrm{rel}}}(\cdot\mid V,z^{\mathrm{irr}})$
        \State $\widehat V_1^{(2)}\gets V_0+D_\psi(V_0,z^{\mathrm{irr}},z^{\mathrm{rel}})$
        \State Update only $E_{\mathrm{rel}}$ using $\mathcal{L}_{\mathrm{rel}}$
    \EndWhile
    \State Freeze $E_{\mathrm{rel}}$
    \Statex \textbf{Conditional rectified-flow dynamics}
    \State Initialize the endpoint predictor $f_\theta$
    \While{not converged}
        \State Sample target clip, preceding history, and instruction $(x_t,h_t,\ell)$
        \State $H_t\gets E_{\mathrm{Wan}}(h_t)$; $e_\ell\gets f_{\mathrm{text}}(\ell)$
        \State Extract $z_t^{\mathrm{rel}}$ from $x_t$ using the frozen encoders and $e_\ell$
        \State Sample $\tau\sim\mathcal{U}(\tau_{\min},1)$ and $\epsilon\sim\mathcal{N}(0,I)$
        \State $z_{\tau,t}\gets(1-\tau)z_t^{\mathrm{rel}}+\tau\epsilon$
        \State Update $f_\theta$ with $\frac{1}{d_z}\|f_\theta(z_{\tau,t},H_t,e_\ell,\tau)-z_t^{\mathrm{rel}}\|_2^2$
    \EndWhile
    \State Freeze $f_\theta$ and compute $(\mu_D,\sigma_D)$ on the pretraining videos
    \end{algorithmic}
\end{algorithm}
\begin{algorithm}[!htbp]
    \caption{Online Policy Learning with DiTaR Rewards}
    \label{alg:suppl_online_rl}
    \begin{algorithmic}[1]
    \footnotesize
    \Require Frozen DiTaR model; task instruction $\ell$; normalization statistics $(\mu_D,\sigma_D)$; environment $\mathcal{E}$
    \Ensure Trained policy $\pi_\omega$
    \State Initialize $\pi_\omega$, replay buffer $\mathcal{B}$, and RND predictor
    \State $e_\ell \gets f_{\mathrm{text}}(\ell)$
    \While{not converged}
        \State Sample $a_t\sim\pi_\omega(\cdot\mid o_t)$, step $\mathcal{E}$, and update the visual history
        \State Form the current and previous reward clips and their preceding histories
        \State Evaluate $e_t$ and $e_{t-\Delta t}$ using Eq.~\eqref{eq:suppl_endpoint_residual}
        \State $r_t^{\mathrm{TD}}\gets\exp(-e_t)-\exp(-e_{t-\Delta t})$
        \State $\widetilde r_t^{\mathrm{TD}}\gets(r_t^{\mathrm{TD}}-\mu_D)/\sigma_D$
        \State Compute $r_t^{\mathrm{RND}}$
        \State $R_t\gets\lambda_D\widetilde r_t^{\mathrm{TD}}+r_t^{\mathrm{env}}+\lambda_{\mathrm{expl}}r_t^{\mathrm{RND}}$
        \State Store the transition with reward $R_t$ in $\mathcal{B}$
        \State Update DrQ-v2 from replay samples and update the RND predictor
    \EndWhile
    \end{algorithmic}
\end{algorithm}

\end{document}